\documentclass{article}

\def \TRtitle{Multi-Task Classification Hypothesis Space with Improved Generalization Bounds}
\def \TRauthors{Cong Li, Michael Georgiopoulos and Georgios C. Anagnostopoulos}
\def \TRemails{congli@eecs.ucf.edu, michaelg@ucf.edu and georgio@fit.edu}
\def \TRaffiliations{CL, MG: EE \& CS Dept., University of Central Florida; GCA: ECE Dept., Florida Institute of Technology}
\def \TRkeywords{Multi-task Learning, Kernel Methods, Generalization Bound, Support Vector Machines}

\usepackage{./sty/options}
\usepackage{./sty/packages}
\usepackage{./sty/macros}
\begin{document}

% Make title pages
\maketitle

% Do not change. %
\ifMakeReviewDraft
	\linenumbers
\fi

\begin{abstract}
 
This paper presents a RKHS, in general, of vector-valued functions intended to be used as hypothesis space for multi-task classification. It extends similar hypothesis spaces that have previously considered in the literature. Assuming this space, an improved Empirical Rademacher Complexity-based generalization bound is derived. The analysis is itself extended to an MKL setting. The connection between the proposed hypothesis space and a Group-Lasso type regularizer is discussed. Finally, experimental results, with some SVM-based Multi-Task Learning problems, underline the quality of the derived bounds and validate the paper's analysis.

\end{abstract}

% Do not change. %

% Input sections from separate files. Modify these fields as necessary.
%%%%%%%%%%%%%%%%%%%%%%%%%%%%%%%%%%%%%%%%%%%%%%%%%%%%%%%%%%%%%%%%%%%%%%%%%%%%%%%%
%%%%%%%%%%%%%%%%%%%%%%%%%%%%%%%%%%%%%%%%%%%%%%%%%%%%%%%%%%%%%%%%%%%%%%%%%%%%%%%%
%%%%%%%%%%%%%%%%%%%%%%%%%%%%%%%%%%%%%%%%%%%%%%%%%%%%%%%%%%%%%%%%%%%%%%%%%%%%%%%%
\section{Introduction}
\label{sec:Introduction}

% Reset all acronyms
\acresetall

\ac{MTL} has been an active research field for over a decade \cite{Caruana1997}. The fundamental philosophy of \ac{MTL} is to simultaneously train several related tasks with shared information, so that the hope is to improve the generalization performance of each task by the assistance of other tasks. More formally, in a typical \ac{MTL} setting with $T$ tasks, we want to choose $T$ functions $f = (f_1, \cdots, f_T)$ from a \ac{HS} $\mathcal{F} = \{  \boldsymbol{x} \mapsto [f_1(\boldsymbol{x}), \cdots, $ $f_T(\boldsymbol{x})]' \}$, where $\boldsymbol{x}$ is an instance of some input set $\mathcal{X}$, such that the performance of each task is optimized based on a problem-specific criterion. Here $[f_1(\boldsymbol{x}), \cdots, f_T(\boldsymbol{x})]'$ denotes the transposition of row vector $[f_1(\boldsymbol{x}), \cdots,f_T(\boldsymbol{x})]$. \ac{MTL} has been successfully applied in feature selection \cite{Argyriou2008,Fei2011,Gong2012}, regression \cite{Lozano2012,Kolar2012}, metric learning \cite{Zhang2010}\cite{Parameswaran2012}, and kernel-based \ac{MTL} \cite{Evgeniou2005,Caponnetto2008,Rakotomamonjy2011,Aflalo2011} among other applications. 

Despite the abundance of \ac{MTL} applications, relevant generalization bounds have only been developed for special cases. A theoretically well-studied \ac{MTL} framework is regularized linear \ac{MTL} model, whose generalization bound is studied in \cite{Maurer2006,Kakade2012,Maurer2012}. In this framework, each function $f_t$ is featured as a linear function with weight $\boldsymbol{w}_t \in \mathcal{H}$, such that $\forall \boldsymbol{x} \in \mathcal{X} \subseteq \mathcal{H}, f_t(\boldsymbol{x}) = \langle \boldsymbol{w}_t, \boldsymbol{x} \rangle$, where $\mathcal{H}$ is a real Hilbert space equipped with inner product $\langle \cdot, \cdot \rangle$. Different regularizers can be employed to the weights $\boldsymbol{w} = (\boldsymbol{w}_1,\cdots, \boldsymbol{w}_T) \in \underbrace{\mathcal{H} \times \ldots \times \mathcal{H}}_{T \text{times}}$ to fulfill different requirements of the problem at hand. Formally, given $\left \{ \boldsymbol{x}_t^i, y_t^i \right \} \in \mathcal{X} \times \mathcal{Y}, i = 1,\cdots,n_t, t = 1,\cdots, T$, where $\mathcal{X}$ and $\mathcal{Y}$ are the input and output space for each task, the framework can be written as

\begin{equation}
\min_{\boldsymbol{w}} R(\boldsymbol{w}) + \lambda \sum_{i,t} L(f_t(\boldsymbol{x}_t^i), y_t^i)
\label{eq:framework_first_category}
\end{equation}

\noindent
where $R(\cdot)$ and $L(\cdot, \cdot)$ are the regularizer and loss function respectively. Many \ac{MTL} models fall into this framework. For example, \cite{Chen2011} looks for group sparsity of $\boldsymbol{w}$, \cite{Zhong2012} discovers group structure of multiple tasks, and \cite{Fei2011,Gong2012} select features in a \ac{MTL} context.

In the previous framework, tasks are implicitly related by regularizers on $\boldsymbol{w}$. On the other hand, another angle of considering information sharing amongst tasks is by pre-processing the data from all tasks by a common processor, and subsequently, a linear model is learned based on the processed data. One scenario of this learning framework is subspace learning, where data of each task are projected to a common subspace by an operator $A$, and then the $\boldsymbol{w}_t$'s are learned in that subspace. Such an approach is followed in \cite{Argyriou2008,Kang2011}. Another particularly straightforward and useful adaptation of this framework is kernel-based \ac{MTL}. In this situation, the role of the operator $A$ is assumed by the non-linear feature mapping $\phi$ associated with the kernel function in use. In this case, all data are pre-processed by a \textit{common} kernel function, which is pre-selected or learned during the training phase, while the $\boldsymbol{w}_t$'s are then learned in the corresponding \ac{RKHS}. One example of this technique is given in \cite{Tang2009}. 

One previous work which discussed the generalization bound of this method in a classification context is \cite{Maurer2006a}. Given a set $\mathcal{A}$ of bounded self-adjoint linear operators on $\mathcal{X}$ and $T$ linear functions with weights $\boldsymbol{w}_t$'s, the \ac{HS} is given as $\mathcal{F} = \{ \boldsymbol{x} \mapsto [ \langle \boldsymbol{w}_1, A\boldsymbol{x} \rangle, \cdots, \langle \boldsymbol{w}_T, A\boldsymbol{x} \rangle]' : \left \| \boldsymbol{w}_t \right \|^2 \leq R, A \in \mathcal{A} \}$. Clearly, in this \ac{HS}, data are pre-processed by the operator $A$ to a common space, as a strategy of information sharing amongst tasks. By either cleverly choosing $A$ beforehand  or by learning $A \in \mathcal{A}$, it is expected that a tighter generalization bound can be attained compared to learning each task independently. It is straightforward to see that pre-selecting $A$ beforehand is a special case of learning $A \in \mathcal{A}$, \ie, pre-selecting $A$ is equivalent to $\mathcal{A} = \{ A \}$.

However, the limitations of $\mathcal{F}$ are two-fold. First, in $\mathcal{F}$, all $\boldsymbol{w}_t$'s are equally constrained in a ball, whose radius $R$ is determined prior to training. However, in practice, the \ac{HS} that lets each task have its own radius for the corresponding norm-ball constraint may be more appropriate and may lead to a better generalization bound and performance. 

%\gcacomment{Why would we expect that? Let's discuss...} 

%\mycomment{If you let each task has its own radius, it will be a more general case compared to equal-radius case. Then the more general one will not perform worse than the special one, right? In the worst case, the special case happens to be the best model, then the more general case will have exactly the same performance as the special one.}

%\gcacomment{You are right once again.}

The second limitation is that it cannot handle the models which learn a common kernel function for all tasks, \eg, the \ac{MT-MKL} models. One way to incorporate such kernel learning models into $\mathcal{F}$ is to let $A$ be the feature mapping $\phi : \mathcal{X} \mapsto \mathcal{H}_{\phi}$, where $\mathcal{H}_\phi$ is the output space of the feature mapping $\phi$ and $\phi$ corresponds to the common kernel function $k$. In other words, this setting defines $\mathcal{F} = \{ \boldsymbol{x} \mapsto [ \langle \boldsymbol{w}_1, \phi(\boldsymbol{x}) \rangle, \cdots, \langle \boldsymbol{w}_T, \phi(\boldsymbol{x}) \rangle]' : \left \| \boldsymbol{w}_t \right \|^2 \leq R, \phi \in \Omega(\phi) \}$, where $\Omega(\phi)$ is the set of feature mappings that $\phi$ is learned from. Obviously, the \ac{HS} that is considered in \cite{Maurer2006a} does not cover this scenario, since it only allows the operator $A$ to be linear operator, which is not the case when $A = \phi$. Yet another limitation reveals itself, when one considers the equivalent \ac{HS}: $\mathcal{F} = \{ \boldsymbol{x} \mapsto [ \langle \boldsymbol{w}_1, \boldsymbol{x} \rangle, \cdots, \langle \boldsymbol{w}_T, \boldsymbol{x} \rangle]' : \left \| \boldsymbol{w}_t \right \|^2 \leq R, \boldsymbol{x} \in \mathcal{H}_\phi, \phi \in \Omega(\phi) \}$, where, as mentioned above, $\mathcal{H}_\phi$ is the output space of the feature mapping $\phi$. Obviously, the \ac{HS} in \cite{Maurer2006a} fails to cover this \ac{HS}, due to the lack of the constraint $\phi \in \Omega(\phi)$, which indicates that the feature mapping $\phi$ (hence, its corresponding kernel function) is learned during the training phase instead of of being selected beforehand.

Therefore, in this paper, we generalize $\mathcal{F}$, particularly for kernel-based classification problems, by considering the common operator $\phi$ (which is associated with a kernel function) for all tasks and by imposing norm-ball constraints on the $\boldsymbol{w}_t$'s with different radii that are learned during the training process, instead of being chosen prior to training. Specifically, we consider the \ac{HS} 

\begin{equation}
\mathcal{F}_s \triangleq  \{ \boldsymbol{x} \mapsto [ \langle \boldsymbol{w}_1, \phi(\boldsymbol{x}) \rangle, \cdots, \langle \boldsymbol{w}_T, \phi(\boldsymbol{x}) \rangle]' : \left \| \boldsymbol{w}_t \right \|^2 \leq \lambda_t^2 R, \boldsymbol{\lambda} \in \Omega_s(\boldsymbol{\lambda})\}
\label{eq:hypothesis_space_intro}
\end{equation}

\noindent
and

\begin{equation}
\mathcal{F}_{s, r} \triangleq \{ \boldsymbol{x} \mapsto [ \langle \boldsymbol{w}_1, \phi(\boldsymbol{x}) \rangle, \cdots, \langle \boldsymbol{w}_T, \phi(\boldsymbol{x}) \rangle]' :  \left \| \boldsymbol{w}_t \right \|^2 \leq \lambda_t^2 R, \boldsymbol{\lambda} \in \Omega_s(\boldsymbol{\lambda}), \phi \in \Omega_r(\phi) \}
\label{eq:hypothesis_space_learn_feature_intro}
\end{equation}

\noindent
where $\Omega_s(\boldsymbol{\lambda}) \triangleq \{ \boldsymbol{\lambda} \succeq \boldsymbol{0}, \left \| \boldsymbol{\lambda} \right \|_s \leq 1, s \geq 1 \}$, and $\Omega_r(\phi) = \{ \phi : \phi =  (\sqrt{\theta_1} \phi_1, \cdots, \sqrt{\theta_M} \phi_M), \boldsymbol{\theta} \succeq \boldsymbol{0}, \left \| \boldsymbol{\theta} \right \|_r \leq 1, r \geq 1\}$, $\phi_m \in \mathcal{H}_m$, $\phi \in \mathcal{H}_1 \times \cdots \times \mathcal{H}_M$. The objective of our paper is to derive and analyze the generalization bounds of these two \ac{HS}. Specifically, the first \ac{HS}, $\mathcal{F}_s$, has fixed feature mapping $\phi$, which is pre-selected, and the second \ac{HS}, $\mathcal{F}_{s, r}$, learns the feature mapping via a \ac{MKL} approach. We refer readers to \cite{Lanckriet2004} and a survey paper \cite{Gonen2011} for details on \ac{MKL}.

Obviously, by letting all $\lambda_t$'s equal $1$, $\mathcal{F}_s$ degrades to the equal-radius \ac{HS}, which is a special case of $\mathcal{F}_s$ when $s \rightarrow +\infty$, as we will show in the sequel. By considering the generalization bound of $\mathcal{F}_s$ based on the \ac{ERC} \cite{Maurer2006}, we first demonstrate that the \ac{ERC} is monotonically increasing with $s$, which implies that the tightest bound is achieved, when $s = 1$. We then provide an upper bound for the \ac{ERC} of $\mathcal{F}_s$, which also monotonically increases with respect to $s$. In the optimal case ($s=1$), we achieve a generalization bound of order $O(\frac{\sqrt{\log T}}{T})$, which decreases relatively fast with increasing $T$. On the other hand, when $s \rightarrow +\infty$, the bound does not decrease with increasing $T$, thus, it is less preferred. 

We then derive the generalization bound for the \ac{HS} $\mathcal{F}_{s, r}$, which still features a bound of order $O(\frac{\sqrt{\log T}}{T})$ when $s = 1$, as in the single-kernel setting. Additionally, if $M$ kernel functions are involved, the bound is of order $O(\sqrt{\log M})$, which has been proved to be the best bound that can be obtained in single-task multi-kernel classification \cite{Cortes2010}. Therefore, the optimal order of the bound is also preserved in the \ac{MT-MKL} case. Note that the proofs of all theoretical results are in the Appendix.

After investigating the generalization bounds, we experimentally show that our bound on the \ac{ERC} matches the real \ac{ERC} very well. Moreover, we propose a \ac{MTL} model based on \acp{SVM} as an example of a classification framework that uses $\mathcal{F}_s$ as its \ac{HS}. It is further extended to an \ac{MT-MKL} setting, whose \ac{HS} becomes $\mathcal{F}_{s,r}$. Experimental results on multi-task classification problems 
%two widely-used multi-task classification problems %data sets
%and two handwritten digits recognition problems 
show the effect of $s$ on the generalization ability of our model. In most situations, the optimal results are indeed achieved, when $s = 1$, which matches our technical analysis. For some other results that are not optimal as expected, when $s = 1$, we provide a justification.

%\begin{equation}
%\left \{ \boldsymbol{x} \mapsto (\boldsymbol{w}_1' (A\boldsymbol{x}), \cdots, \boldsymbol{w}_T' (A\boldsymbol{x}))' : \left \| \boldsymbol{w}_t \right \| \leq R, A \in \mathcal{A} \right \}
%\label{eq:previous_hypothesis_space}
%\end{equation}

%%%%%%%%%%%%%%%%%%%%%%%%%%%%%%%%%%%%%%%%%%%%%%%%%%%%%%%%%%%%%%%%%%%%%%%%%%%%%%%%
%%%%%%%%%%%%%%%%%%%%%%%%%%%%%%%%%%%%%%%%%%%%%%%%%%%%%%%%%%%%%%%%%%%%%%%%%%%%%%%%
%%%%%%%%%%%%%%%%%%%%%%%%%%%%%%%%%%%%%%%%%%%%%%%%%%%%%%%%%%%%%%%%%%%%%%%%%%%%%%%%
\section{Fixed Feature Mapping}
\label{sec:FixedFeatureMapping}

%We start with the introduction of problem settings.  
Let $\left \{ \boldsymbol{x}_t^i, y_t^i \right \} \in \mathcal{X} \times \left \{ -1, 1 \right \}, i = 1,\cdots,N, t = 1,\cdots,T$ be i.i.d. training samples from some joint distribution. Without loss of generality and on grounds of convenience, we will assume an equal number of training samples for each task. Let $\mathcal{H}$ be a \ac{RKHS} with reproducing kernel function  $k(\cdot, \cdot) : \mathcal{X} \times \mathcal{X} \mapsto \mathbb{R}$, and associated feature mapping $\phi: \mathcal{X} \mapsto \mathcal{H}$. 
In what follows we give the theoretical analysis of our \ac{HS} $\mathcal{F}_s$, when the feature mapping $\phi$ is fixed.

%%%%%%%%%%%%%%%%%%%%%%%%%%%%%%%%%%%%%%%%%%%%%%%%%%%%%%%%%%%%%
%%%%%%%%%%%%%%%%%%%%%%%%%%%%%%%%%%%%%%%%%%%%%%%%%%%%%%%%%%%%%
\subsection{Theoretical Results}
\label{sec:TheoreticalResultsFixFeature}
Given $T$ tasks, our objective is to learn $T$ linear functionals $f_t(\cdot) : \mathcal{H} \mapsto \mathbb{R}$, such that $f_t(\phi (\boldsymbol{x})) = \left \langle \boldsymbol{w}_t, \phi(\boldsymbol{x}) \right \rangle, t = 1,\cdots,T$, $\boldsymbol{x} \in \mathcal{X}$. Next, let $\boldsymbol{f} \triangleq [f_1, \cdots, f_T]'$, and define the multi-task classification error as 

%\gcacomment{previous: check the correctness of my use of terminology: ``multi-task classification error''; you had ``expected error''}

\begin{equation}
er(f) \triangleq \frac{1}{T} \sum_{t=1}^T E\{\boldsymbol{1}_{(-\infty, 0]} (y_t f_t(\phi (\boldsymbol{x}_t)))\}
\label{eq:expected_error}
\end{equation}

\noindent
where $\boldsymbol{1}_{(-\infty, 0]}(\cdot)$ is the characteristic function of $(-\infty, 0]$ and referred to as the $0/1$ loss function. The empirical error based on a surrogate loss function $\bar{L} : \mathbb{R} \mapsto [0, 1]$, which is a Lipschitz-continuous function that upper-bounds the $0/1$ loss function, is defined as

%\gcacomment{I am unsure if ``surrogate'' loss function is standard terminology; in case it is not, it should be defined: Here's my take: a Lipschitz-continuous function that upper-bounds the 0/1 loss function and takes values in $[0,1]$ (assuming that the reader knows that the 0/1 loss function is $1_{(-\infty,0]}(\cdot)$). On second look, I would rewrite it as: where $\boldsymbol{1}_{(-\infty, 0]}(\cdot)$ is the characteristic function of $(-\infty, 0]$ and referred to as the 0/1 loss function. The empirical error based on a surrogate loss function $L : \mathbb{R} \mapsto [0, 1]$, which is a Lipschitz-continuous function that upper-bounds the 0/1 loss function, is defined as...\\ \\ Important: in lemma 1, you must not forget to say that L upper-bounds the 0/1 loss, otherwise the inequality you mention in lemma 1 is not technically correct. See also if you have omitted it elsewhere... }
%\mycomment{Revised. Thanks.}

\begin{equation}
\hat{er}(f) \triangleq \frac{1}{TN} \sum_{t,i=1}^{T,N} \bar{L}(y_t^i f_t(\phi (\boldsymbol{x}_t^i)))
\label{eq:empirical_error}
\end{equation}

\noindent
For the constraints on the $\boldsymbol{w}_t$'s, instead of pre-defining a common radius $R$ for all tasks as discussed in \cite{Maurer2006a}, we let $\left \| \boldsymbol{w}_t \right \|^2 \leq \lambda_t^2 R$, where $\lambda_t$ is learned during the training phase. This motivates our consideration of $\mathcal{F}_s$ as given in (\ref{eq:hypothesis_space_intro}), which we repeat here:

\begin{equation}
\mathcal{F}_s \triangleq  \{ \boldsymbol{x} \mapsto [ \langle \boldsymbol{w}_1, \phi(\boldsymbol{x}) \rangle, \cdots, \langle \boldsymbol{w}_T, \phi(\boldsymbol{x}) \rangle]' : \left \| \boldsymbol{w}_t \right \|^2 \leq \lambda_t^2 R, \boldsymbol{\lambda} \in \Omega_s(\boldsymbol{\lambda})\}
\label{eq:hypothesis_space}
\end{equation}

\noindent
Note that the feature mapping $\phi$ is determined before training. In order to derive the generalization bound for $\mathcal{F}_s$, we first provide the following lemma.

%(\ref{eq:hypothesis_space}) is equivalent to

%\begin{equation}
%\begin{aligned}
%\mathcal{F}_s \triangleq  & \{ \boldsymbol{x} \mapsto (\lambda_1 \left \langle \boldsymbol{v}_1, \phi(\boldsymbol{x}) \right \rangle, \cdots, \lambda_T \left \langle \boldsymbol{v}_T, \phi(\boldsymbol{x}) \right \rangle)' : \\
% & \left \| \boldsymbol{v}_t \right \|^2 \leq R, \boldsymbol{\lambda} \in \Omega_s(\boldsymbol{\lambda}) \}
%\end{aligned}
%\label{eq:hypothesis_space_equivalent}
%\end{equation}

%\noindent
%if $\boldsymbol{v}_t \triangleq  \frac{\boldsymbol{w}_t}{\lambda_t}$. In other words, searching for functions $f_t$'s with different norm-ball constraint radii (parametrized by the $\lambda_t$'s) is equivalent to finding functions with equal-radii constraint,s but each function is pre-multiplied by a weight $\lambda_t$. 

%\gcacomment{The previous observation is rather trivial; I suggest you remove it.}

%\gcacomment{Next lemma: should it be ``Lipschitz-continuous function'' instead?} \mycomment{Maurer used ``Lipschitz function'', so I became a copycat. However, after some google search, I think these two concepts are equivalent. See \url{http://mathworld.wolfram.com/LipschitzFunction.html}} \gcacomment{copy(*miao*)... Let's go with ``Lipschitz-continuous''...}

%%%%%%%%%%%%%%%%%%%%%%%%%%%%%%%%%%%%%%%%%%%%%%%%%%%%%%%%%%%%%
\begin{lemma}

Let $\mathcal{F}_s$ be as defined in \eref{eq:hypothesis_space}. Let $\bar{L} : \mathbb{R} \mapsto [0, 1]$ be a Lipschitz-continuous loss function with Lipschitz constant $\gamma$ and upper-bound the $0/1$ loss function $\boldsymbol{1}_{(-\infty, 0]}(\cdot)$. Then, with probability $1-\delta$ we have 

\begin{equation}
er(f) \leq \hat{er}(f) + \frac{1}{\gamma} \hat{R}(\mathcal{F}_s) + \sqrt{\frac{9 \log \frac{2}{\delta}}{2TN}} , \;\;\forall f \in \mathcal{F}_s
\label{eq:generalization_bound}
\end{equation}

%\gcacomment{What is $\boldsymbol{f}$? Not defined. Please check the entire manuscript for undefined quantities. \\ \\ Suggestion: Also, I am not too crazy about your use of bold face fonts. It is customary that (i) if x is from an arbitrary input set, it shouldn't be bold; (ii) same thing for elements of an unspecified Hilbert space. All this, unless you are talking about $\mathbb{R}^D.$}
%
%\mycomment{Revised. Thanks.}

\noindent where $\hat{R}(\mathcal{F}_s)$ is the \ac{ERC} for \ac{MTL} problems defined in \cite{Maurer2006}:

\begin{equation}
\hat{R}(\mathcal{F}_s) \triangleq E_{\sigma}\{\sup_{\boldsymbol{f}\in \mathcal{F}_s} \frac{2}{TN} \sum_{t,i=1}^{TN}\sigma_t^i f_t(\phi(\boldsymbol{x}_t^i))\}
\label{eq:rademacher}
\end{equation}

\noindent where the $\sigma_t^i$'s are i.i.d. Rademacher-distributed (\ie, $\mbox{Bernoulli}\left(\frac{1}{2}\right)$-distributed random variables with sample space $\left\{-1, +1\right\}$).

%\gcacomment{Well, these RVs are called Rademacher RVs. You could say something like: \\ where the $\sigma_t^i$'s are i.i.d. Rademacher-distributed (\ie, $\mbox{Bernoulli}\left(\frac{1}{2}\right)$-distributed random variables with sample space $\left\{-1, +1\right\}$).}
%\mycomment{Revised}.

\label{lemma:generalization_bound}
\end{lemma}
%%%%%%%%%%%%%%%%%%%%%%%%%%%%%%%%%%%%%%%%%%%%%%%%%%%%%%%%%%%%%

%\gcacomment{This comment is irrelevant to the present context; it is general: Why talk about Fs and Fs,r separately, since Fs is a subset of Fs,r? Why not consider only Fs,r? What am I missing? }
%\mycomment{I think it will be more logically smooth to first propose the MTL HS only. This is because that, our HS is a generalization of Maurer's (which is MTL HS), so it will take people less effort to accept the MTL-only HS, and then extend it to the MT-MKL case. If we talk about MT-MKL HS at the beginning, and derive the complicated bound, it will be too much info to absorb. But you are right, Fs is a subset of Fs,r.}

%\gcacomment{The ERC should be part of (inside) the lemma.} \mycomment{The thing is, \ac{ERC} for \ac{MTL} problems is firstly given in Maurer2006b. But the lemma is not given in that paper. Strictly speaking, this lemma is proven by me. However, it is simple use of the two theorems that I mentioned below. Thus I will not say it's my contribution. Since the lemma and the ERC are coming from different sources, I put the \ac{ERC} outside of the lemma, otherwise I need to cite Maurer2006b inside the lemma (right beside where I introduce the \ac{ERC}).} \gcacomment{Please put the ERC in the same lemma for completeness and then mention that it comes from two sources.}

\noindent
This lemma can be simply proved by utilizing Theorem 16 and 17 in \cite{Maurer2006a}. By using the same proving strategy, it is easy to show that (\ref{eq:generalization_bound}) is valid for all \acp{HS} that are considered in this paper. Therefore, we will not explicitly state a specialization of it for each additional \ac{HS} encountered in the sequel. In the next, we first define the following duality mapping for all $a \in \mathbb{R}$:

\begin{equation}
(\cdot)^* : a \mapsto a^* \triangleq \left\{\begin{matrix}
\frac{a}{a-1}, & \forall a \neq 1\\ 
+\infty , & a = 1
\end{matrix}\right.
\label{eq:duality}
\end{equation}

%\mycomment{Dr. A, I'm not sure if the statement ``$+\infty \;\text{or} -\infty$'' is appropriate.}

\noindent
then we give the following results which show that $\hat{R}(\mathcal{F}_s)$ is monotonically increasing with respect to $s$.

%%%%%%%%%%%%%%%%%%%%%%%%%%%%%%%%%%%%%%%%%%%%%%%%%%%%%%%%%%%%%
\begin{lemma}

Let $\boldsymbol{\sigma}_t \triangleq [\sigma_t^1,\cdots,\sigma_t^N]'$, $u_t \triangleq \sqrt{\boldsymbol{\sigma}_t^{'} \boldsymbol{K}_t \boldsymbol{\sigma}_t}$, where $\boldsymbol{K}_t$ is the kernel matrix that consists of elements $k(\boldsymbol{x}_t^i, \boldsymbol{x}_t^j), t = 1,\cdots,T$, $\boldsymbol{u} \triangleq [u_1, \cdots, u_T]'$. Then $\forall s \geq 1$

\begin{equation}
\hat{R}(\mathcal{F}_s) = \frac{2}{TN} \sqrt{R} E_{\sigma} \{\left \| \boldsymbol{u} \right \|_{s^*}\}
\label{eq:rademacher_alternative}
\end{equation}

\label{lemma:rademacher_alternative_fix_feature}
\end{lemma}
%%%%%%%%%%%%%%%%%%%%%%%%%%%%%%%%%%%%%%%%%%%%%%%%%%%%%%%%%%%%%

%\gcacomment{cosmetic suggestion: you may consider using $\mathbb{E}\left\{ \cdot \right\}$ for expectations. I would recommend using curly brackets with $\mathbb{E}$, as it is an operator, not a function.}

%\gcacomment{How does the previous lemma connect to the next theorem? You should say something like: ``Based on the previous lemma, one can show the following theorem'' or something like that.} \mycomment{See how it looks.}

\noindent
Leveraging from \eref{eq:rademacher_alternative}, one can show the following theorem.

%%%%%%%%%%%%%%%%%%%%%%%%%%%%%%%%%%%%%%%%%%%%%%%%%%%%%%%%%%%%%
\begin{theorem}
$\hat{R}(\mathcal{F}_s)$ is monotonically increasing with respect to $s$.
\label{thm:rademacher_monotonicity_fix_feature}
\end{theorem}
%%%%%%%%%%%%%%%%%%%%%%%%%%%%%%%%%%%%%%%%%%%%%%%%%%%%%%%%%%%%%

Define $\tilde{\mathcal{F}} \triangleq \{ \boldsymbol{x} \mapsto [ \langle \boldsymbol{w}_1, \phi(\boldsymbol{x}) \rangle, \cdots, \langle \boldsymbol{w}_T, \phi(\boldsymbol{x}) \rangle ]' : \left \| \boldsymbol{w}_t \right \|^2 \leq R \}$, which is the \ac{HS} that is given in \cite{Maurer2006a} under kernelized \ac{MTL} setting, then $\tilde{\mathcal{F}}$ is the \ac{HS} with equal radius for each $\| \boldsymbol{w}_t \|^2$. Obviously, it is the special case of $\mathcal{F}_s$ with all $\lambda_t$'s be set to $1$. We have the following result:

%%%%%%%%%%%%%%%%%%%%%%%%%%%%%%%%%%%%%%%%%%%%%%%%%%%%%%%%%%%%%
\begin{theorem}
$\hat{R}(\tilde{\mathcal{F}}) = \hat{R}(\mathcal{F}_{+\infty})$.
\label{thm:s_infty_fix_feature}
\end{theorem}
%%%%%%%%%%%%%%%%%%%%%%%%%%%%%%%%%%%%%%%%%%%%%%%%%%%%%%%%%%%%%

%\gcacomment{Several comments: (i) So equality of complexities is equivalent to one HS being a special case of another? It seems to me like the theorem's statement is badly phrased. I think what matters is that the two spaces have the same ERC; that's it. If you had defined your spaces such that $|lambda|_s <= Ulambda$ (you use Ulambda=1), and Ulambda is big enough for the given s, then indeed Maurer's HS is included in yours. BTW, why did you fix the constraint bounds both for lambda and theta to 1? Why not have them Ulambda and Utheta and then just set them equal to 1 for your experiments? Finally, what's the usefulness of the previous theorem?}
%
%\mycomment{1. You are right about the first question, we only need to state the ERC of the two HSs are the same. See how the theorem looks now. \\ \\ 2. Regarding the Ulambda issue, I think you have some misunderstanding here. Let's talk when I get back to school. \\ \\ 3. The usefulness of the theorem is to say that, the equal radii HS, which is given in Maurer's paper, is the worst choice. I revised the paragraph above and below the theorem, please see if the point is more clear now.}
%
%\gcacomment{1. I would just state the theorem simply as ``$\hat{R}(\tilde{\mathcal{F}}) = \hat{R}(\mathcal{F}_{+\infty})$''; wouldn't use ``i.e.''}
%
%\gcacomment{2. That would be of high interest to me, as I am apparently missing an important point here.}
%
%\gcacomment{3. Great.}

The above results imply that the tightest generalization bound is obtained when $s=1$, while, on the other hand, the bound of $\mathcal{F}_{+\infty}$ that sets equal radii for all $\boldsymbol{w}_t$'s is the least preferred. It is clear that, to derive a generalization bound for $\mathcal{F}_s$, we need to compute, or, at least find an upper bound for $\hat{R}(\mathcal{F}_s)$. The following theorem addresses this requisite.

%\gcacomment{Below: Do not use ``for $\forall$''... just ``$\forall$''}

%\gcacomment{Please reconsider your notation: it probably would make sense to exchange the names of $\hat{s}^*$ and $\hat{s}$ to preserve the notation of dual.} \mycomment{True. But if I change notations, when I write proof by copying-pasting from my notes, it will be completely messed up. In case this happens, I removed the terminology ``dual'', since I didn't use it in Theorem 1 neither.}

%\gcacomment{I am OK with that; besides, it is your paper. However, I am of the impression that you under-appreciate the value of good notation, probably because you see what others do (or don't do). That does not mean it should be this way. For example, others may not care (and some times, even do not want) if their audience can/should easily understand the work presented. }

%%%%%%%%%%%%%%%%%%%%%%%%%%%%%%%%%%%%%%%%%%%%%%%%%%%%%%%%%%%%%
\begin{theorem}

Let $\mathcal{F}_s$ be as defined in \eref{eq:hypothesis_space}, and let $\rho \triangleq 2 \ln T$. Assume that $\forall \boldsymbol{x} \in \mathcal{X}$, $k(\boldsymbol{x}, \boldsymbol{x}) = \left \langle \phi(\boldsymbol{x}), \phi(\boldsymbol{x}) \right \rangle \leq 1$. Then the \ac{ERC} can be bounded as follows:

\begin{equation}
\hat{R}(\mathcal{F}_s) \leq \frac{2}{T\sqrt{N}} \sqrt{  \tau  RT^{\frac{2}{s^*}} }
\label{eq:rademacher_fix_feature}
\end{equation}

\noindent
where $\tau \triangleq \left( \max \left\{ s, \rho^* \right\} \right)^*$.
%\noindent
%where
%
%\begin{equation}
%\tilde{s} \triangleq \left\{\begin{matrix}
%s^* & \text{if} \; s > \hat{s} \\
%\hat{s}^* & \text{if} \; s \leq \hat{s}
%\end{matrix}\right.
%\label{eq:sTilde_fix_feature}
%\end{equation}

\label{thm:rademacher_fix_feature}
\end{theorem}
%%%%%%%%%%%%%%%%%%%%%%%%%%%%%%%%%%%%%%%%%%%%%%%%%%%%%%%%%%%%%

%%%%%%%%%%%%%%%%%%%%%%%%%%%%%%%%%%%%%%%%%%%%%%%%%%%%%%%%%%%%%
%%%%%%%%%%%%%%%%%%%%%%%%%%%%%%%%%%%%%%%%%%%%%%%%%%%%%%%%%%%%%
\subsection{Analysis}
\label{sec:FixedFeatureAnalysis}

It is worth pointing out some observations regarding the result of \thmref{thm:rademacher_fix_feature}.

\begin{itemize}
	
	\item It is not difficult to see that the bound of the \ac{ERC} in (\ref{eq:rademacher_fix_feature}) is monotonically increasing in $s$, as is $\hat{R}(\mathcal{F}_s)$. 
	
	\item As $s \rightarrow +\infty$, $\mathcal{F}_s$ degrades to $\tilde{\mathcal{F}}$. In this case, $\hat{R}(\mathcal{F}_{+\infty}) \leq 2 \sqrt{\frac{R}{N}}$. Note that this bound matches the one that is given in \cite{Maurer2006a}. This is because of the following relation between $\tilde{\mathcal{F}}$ and the \ac{HS} of \cite{Maurer2006a}, $\mathcal{F}$, that is introduced in \sref{sec:Introduction}: First, let the operator $A$ in $\mathcal{F}$ be the identity operator, and then let $\boldsymbol{x}$ in $\mathcal{F}$ be an element of $\mathcal{H}$, \ie, let $\boldsymbol{x}$ in $\mathcal{F}$ be $\phi(\boldsymbol{x})$ in $\tilde{\mathcal{F}}$. Then $\mathcal{F}$ becomes $\tilde{\mathcal{F}}$.
	
	%\gcacomment{Is it indeed worth mentioning? If Ftilde is a subset of F and F is a subset of Ftilde, then they the same, which, obviously, is not the case after eyeballing their definitions. I suggest, remove the previous sentence.} 
	%\mycomment{Right. Deleted.}
		
	\item Obviously, when $s$ is finite, the bound for $\mathcal{F}_s$, which is of order $O(\frac{1}{T^{\frac{1}{s}}} \sqrt{\frac{1}{N}})$, is more preferred over the aforementioned $O(\frac{1}{\sqrt{N}})$ bound, as it asymptotically decreases with increasing number of tasks.  
	%Obviously, compare to its bound $O(\frac{1}{\sqrt{N}})$ , the bound of $\mathcal{F}_s$ with finite $s$, which is $O(\frac{1}{T^{\frac{1}{s}}} \sqrt{\frac{\tilde{s}}{N}})$, is more preferred, as it asymptotically decreases with increasing number of tasks. 
	
	%\gcacomment{Actually, if you care only for N \& T, your bound is $O\left( \frac{1}{ T^{\frac{1}{s}} \sqrt{N}} \right)$, since your $\tilde{s} = (\max(\cdot))^* = O(1)$ (bounded) in terms of T. } 
	%\mycomment{True. Revised.}
	
	\item When $s = \rho^*$, $\hat{R}(\mathcal{F}_{\rho^*}) \leq \frac{2}{T\sqrt{N}} \sqrt{2eR \log T}$. Here we achieve a bound of order $O(\frac{\sqrt{\log T}}{T})$, which decreases faster with increasing $T$ compared to the bound, when $s > \rho^*$. 
	
	%\gcacomment{Previous: Checked; correct.}
	
	\item When $s = 1$, $\hat{R}(\mathcal{F}_1) \leq \frac{2}{T\sqrt{N}} \sqrt{2R \log T}$. While being of order $O(\frac{\sqrt{\log T}}{T})$, it features a smaller constant compared to the bound of $\hat{R}(\mathcal{F}_{\rho^*})$. In fact, due to the monotonicity of the bound that is given in (\ref{eq:rademacher_fix_feature}), the tightest bound is obtained when $s = 1$.

%\gcacomment{If what I mentioned earlier is correct, the comparison statement may be wrong and must be scrutinized.}
	
\end{itemize}

%It is worth mentioning that our result extends Corollary 10 in \cite{Kloft2011} and Proposition 2 in \cite{Kloft2012} to a \ac{MTL} setting; our bound is tighter by a constant factor $e$.

In the next section, we derive and analyze the generalization bound by letting $\phi$ to be learned during the training phase. 
%We show that similar results can be obtained: the \ac{HS} with fixed radii of norm-ball constraints yields loser bounds compared to our proposed \ac{HS}.

%%%%%%%%%%%%%%%%%%%%%%%%%%%%%%%%%%%%%%%%%%%%%%%%%%%%%%%%%%%%%%%%%%%%%%%%%%%%%%%%
%%%%%%%%%%%%%%%%%%%%%%%%%%%%%%%%%%%%%%%%%%%%%%%%%%%%%%%%%%%%%%%%%%%%%%%%%%%%%%%%
%%%%%%%%%%%%%%%%%%%%%%%%%%%%%%%%%%%%%%%%%%%%%%%%%%%%%%%%%%%%%%%%%%%%%%%%%%%%%%%%
\section{Learning the Feature Mapping}
\label{sec:LearningFeatureMapping}

In this section, we consider the selection of the feature mapping $\phi$ during training via an \ac{MKL} approach. In particular, we will assume that $\phi = (\sqrt{\theta_1} \phi_1, \cdots, \sqrt{\theta_M} \phi_M) \in \mathcal{H}_1 \times \cdots \times \mathcal{H}_M$, where each $\phi_m : \mathcal{X} \mapsto \mathcal{H}_m$ is selected before training.

%%%%%%%%%%%%%%%%%%%%%%%%%%%%%%%%%%%%%%%%%%%%%%%%%%%%%%%%%%%%%%%%%%%%%%%%%
%%%%%%%%%%%%%%%%%%%%%%%%%%%%%%%%%%%%%%%%%%%%%%%%%%%%%%%%%%%%%%%%%%%%%%%%%
\subsection{Theoretical Results}
\label{sec:TheoreticalResultsLearnFeature}

Consider the following \ac{HS}

\begin{equation}
\mathcal{F}_{s, r} \triangleq  \{ \boldsymbol{x} \mapsto [ \langle \boldsymbol{w}_1, \phi(\boldsymbol{x}) \rangle, \cdots, \langle \boldsymbol{w}_T, \phi(\boldsymbol{x}) \rangle ]' :  \left \| \boldsymbol{w}_t \right \|^2 \leq \lambda_t^2 R, \boldsymbol{\lambda} \in \Omega_s(\boldsymbol{\lambda}), \phi \in \Omega_r(\phi) \}
\label{eq:hypothesis_space_learn_feature}
\end{equation}

\noindent
where $\Omega_r(\phi) = \{ \phi : \phi =  (\sqrt{\theta_1} \phi_1, \cdots, \sqrt{\theta_M} \phi_M), \boldsymbol{\theta} \succeq \boldsymbol{0}, \left \| \boldsymbol{\theta} \right \|_r \leq 1\}$. By following the same derivation procedure of \lemmaref{lemma:generalization_bound}, we can verify that (\ref{eq:generalization_bound}) is also valid for $\mathcal{F}_{s, r}$. Therefore, we only need to estimate its \ac{ERC}. Similar to the previous section, we first give results regarding the monotonicity of $\hat{R}(\mathcal{F}_{s, r})$.

%%%%%%%%%%%%%%%%%%%%%%%%%%%%%%%%%%%%%%%%%%%%%%%%%%%%%%%%%%%%%
\begin{lemma}

Let  $\boldsymbol{\sigma}_t \triangleq [\sigma_t^1,\cdots,\sigma_t^N]'$, $u_t^m \triangleq \boldsymbol{\sigma}_t^{'} \boldsymbol{K}_t^m \boldsymbol{\sigma}_t$, $\boldsymbol{u}_t \triangleq [u_t^1,\cdots,u_t^M]'$, $v_m \triangleq \sum_{t=1}^T \lambda_t \boldsymbol{\sigma}_t^{'} \boldsymbol{K}_t^m \boldsymbol{\alpha}_t$, $\boldsymbol{v} \triangleq [v_1, \cdots, v_M]'$, where $\boldsymbol{K}_t^m$ is the kernel matrix that contains elements $k_m(\boldsymbol{x}_t^i, \boldsymbol{x}_t^j)$. Then $\forall s \geq 1$ and $r \geq 1$,

\begin{equation}
\hat{R}(\mathcal{F}_{s, r}) = \frac{2}{TN} \sqrt{R} E_{\sigma} \{\sup_{\boldsymbol{\theta} \in \Omega_r(\boldsymbol{\theta})} \sum_{t=1}^T (\boldsymbol{\theta}' \boldsymbol{u}_t)^{\frac{s^*}{2}}\}^{\frac{1}{s^*}} = \frac{2}{TN} E_{\sigma} \{\sup_{\boldsymbol{\lambda} \in \Omega_s(\boldsymbol{\lambda}), \boldsymbol{\alpha} \in \Omega(\boldsymbol{\alpha})} \left \| \boldsymbol{v} \right \|_{r^*}\}
\label{eq:rademacher_alternative_learn_feature_1}
\end{equation}

%\begin{equation}
%\hat{R}(\mathcal{F}_{s, r}) = \frac{2}{TN} E_{\sigma} \{\sup_{\boldsymbol{\lambda} \in \Omega_s(\boldsymbol{\lambda}), \boldsymbol{\alpha} \in \Omega(\boldsymbol{\alpha})} \left \| \boldsymbol{v} \right \|_{r^*}\}
%\label{eq:rademacher_alternative_learn_feature_2}
%\end{equation}

%\gcacomment{Why do you provide two alternate expressions here? Also, define vector transposition and remove the re-definitions of starred quantities, if you liked what I wrote earlier.}
%
%\mycomment{(14) is used to prove Theorem 4, and (13) is for Theorem 5. \\ The vector transposition is defined at the beginning of the paper.}
%
%\gcacomment{I am still not happy. Present only one; move the other one (maybe, along with its derivation, into the corresponding theorem. Alternative: combine both into one equation line e.g. R = expression1 = expression2.}

\noindent
where $\Omega_r(\boldsymbol{\theta}) \triangleq \{ \boldsymbol{\theta} : \boldsymbol{\theta} \succeq \boldsymbol{0}, \left \| \boldsymbol{\theta}\right \|_r \leq 1 \}$, and $\Omega(\boldsymbol{\alpha}) \triangleq \{ \boldsymbol{\alpha}_t : \boldsymbol{\sigma}_t^{'} \boldsymbol{K}_t^m \boldsymbol{\alpha}_t \leq R, \forall t \}$.

\label{lemma:rademacher_alternative_learn_feature}
\end{lemma}
%%%%%%%%%%%%%%%%%%%%%%%%%%%%%%%%%%%%%%%%%%%%%%%%%%%%%%%%%%%%%

\noindent
Based on \eref{eq:rademacher_alternative_learn_feature_1}, we have the following result:

%%%%%%%%%%%%%%%%%%%%%%%%%%%%%%%%%%%%%%%%%%%%%%%%%%%%%%%%%%%%%
\begin{theorem}
$\hat{R}(\mathcal{F}_{s, r})$ is monotonically increasing with respect to $s$.
\label{thm:rademacher_monotonicity_learn_feature}
\end{theorem}
%%%%%%%%%%%%%%%%%%%%%%%%%%%%%%%%%%%%%%%%%%%%%%%%%%%%%%%%%%%%%

We extend $\tilde{\mathcal{F}}$ to a \ac{MT-MKL} setting by letting $\phi \in \Omega_r(\phi)$, for $\phi$ in $\tilde{\mathcal{F}}$, which gives $\tilde{\mathcal{F}}_r \triangleq \{ \boldsymbol{x} \mapsto (\left \langle \boldsymbol{w}_1, \phi(\boldsymbol{x}) \right \rangle, \cdots, \left \langle \boldsymbol{w}_T, \phi(\boldsymbol{x}) \right \rangle)' : \left \| \boldsymbol{w}_t \right \|^2 \leq R, \phi \in \Omega_r(\phi) \}$. Then, \eref{eq:rademacher_alternative_learn_feature_1} leads to the following result:

%%%%%%%%%%%%%%%%%%%%%%%%%%%%%%%%%%%%%%%%%%%%%%%%%%%%%%%%%%%%%
\begin{theorem}
$\hat{R}(\tilde{\mathcal{F}}_r) = \hat{R}(\mathcal{F}_{+\infty, r})$ and, thus, $\tilde{\mathcal{F}_r}$ is a special case of $\mathcal{F}_{s, r}$.
\label{thm:s_infty_learn_feature}
\end{theorem}
%%%%%%%%%%%%%%%%%%%%%%%%%%%%%%%%%%%%%%%%%%%%%%%%%%%%%%%%%%%%%

Again, the above results imply that the tightest bound is obtained, when $s = 1$. In the following theorem, we provide an upper bound for $\hat{R}(\mathcal{F}_{s, r})$.

%%%%%%%%%%%%%%%%%%%%%%%%%%%%%%%%%%%%%%%%%%%%%%%%%%%%%%%%%%%%%
\begin{theorem}

Let $\mathcal{F}_{s, r}$ be as defined in \eref{eq:hypothesis_space_learn_feature}. Assume that $\forall \boldsymbol{x} \in \mathcal{X}$, $m = 1,\cdots,M$, $k_m(\boldsymbol{x}, \boldsymbol{x}) = \left \langle \phi_m(\boldsymbol{x}), \phi_m(\boldsymbol{x}) \right \rangle \leq 1$. The \ac{ERC} can be bounded as follows:

%\gcacomment{It's a shame that you did not parameterized the max value of the kernels to obtain a more general formula with a little bit of extra effort.}

%\mycomment{I would say it's not necessary. I'll explain to you when I go back.}

%\gcacomment{OK, sounds good.}

\begin{equation}
\hat{R}(\mathcal{F}_{s, r}) \leq \frac{2}{T\sqrt{N}} \sqrt{R s^* T^{\frac{2}{s^*}} M^{\max \{ \frac{1}{r^*}, \frac{2}{s^*} \}}}
\label{eq:rademacher_learn_feature}
\end{equation}

\label{thm:rademacher_learn_feature}
\end{theorem}
%%%%%%%%%%%%%%%%%%%%%%%%%%%%%%%%%%%%%%%%%%%%%%%%%%%%%%%%%%%%%

The above theorem can be explicitly refined under the following two situations:

%%%%%%%%%%%%%%%%%%%%%%%%%%%%%%%%%%%%%%%%%%%%%%%%%%%%%%%%%%%%%
\begin{corollary}
Under the conditions that are given in \thmref{thm:rademacher_learn_feature}, we have

\begin{equation}
\begin{aligned}
\hat{R}(\mathcal{F}_{s, r}) & \leq \frac{2}{T\sqrt{N}} \sqrt{\tau R T^{\frac{2}{s^*}} M^{\frac{1}{r^*}}} \;\;\;\; & \text{if} \; r^* \leq \log T \\
\hat{R}(\mathcal{F}_{s, r}) & \leq \frac{2}{T\sqrt{N}} \sqrt{\tau R T^{\frac{2}{s^*}} M^{\frac{2}{\tau}}} \;\;\;\; & \text{if} \; r^* \geq \log MT
\end{aligned}
\label{eq:rademacher_learn_feature_r1}
\end{equation}

\noindent
$\forall s \geq 1$, where $\tau \triangleq \left( \max \left\{ s, \rho^* \right\} \right)^*$, and

\begin{equation}
\rho \triangleq \left\{\begin{matrix}
2 \ln T, & r^* \leq \ln T\\ 
2 \ln MT, & r^* \geq \ln MT
\end{matrix}\right.
\label{eq:rho_learn_feature}
\end{equation}

\label{col:rademacher_learn_feature_r}
\end{corollary}
%%%%%%%%%%%%%%%%%%%%%%%%%%%%%%%%%%%%%%%%%%%%%%%%%%%%%%%%%%%%%

%\gcacomment{From (18), the range of $r^*$ from $\log T$ to $\log T + \log M$ is missing... }

%\mycomment{Actually in this range, the result is given in Theorem 6. Theorem 6 is a universally correct result, but not tight for the two ranges that are given in this corollary.}

%%%%%%%%%%%%%%%%%%%%%%%%%%%%%%%%%%%%%%%%%%%%%%%%%%%%%%%%%%%%%%%%%%%%%%%%%
%%%%%%%%%%%%%%%%%%%%%%%%%%%%%%%%%%%%%%%%%%%%%%%%%%%%%%%%%%%%%%%%%%%%%%%%%
\subsection{Analysis}
\label{sec:AnalysisLearnFeature}
Once again, it is worth commenting on the results given in \thmref{thm:rademacher_learn_feature} and \colref{col:rademacher_learn_feature_r}:

%\gcacomment{I did not scrutinize the following analysis as I did the previous one in the interest of time.}

\begin{itemize}

	\item Generally, $\forall s \geq 1$, (\ref{eq:rademacher_learn_feature_r1}) gives a bound of order $O(\frac{1}{T^{\frac{1}{s}}})$. Obviously, $s \rightarrow +\infty$ is least preferred, since its bound does not decrease with increasing number of tasks. Moreover, based on (\ref{eq:rademacher_learn_feature}), $\forall r \geq 1$, $\hat{R}(\mathcal{F}_{1, r})$'s bound is of order $O(\sqrt{M^{\frac{1}{r^*}}})$. Compared to the $O(\sqrt{M^{\frac{1}{r^*}} \min(\lceil \log M \rceil, \lceil r^* \rceil)})$ bound of single-task \ac{MKL} scenario, which is examined in \cite{Kloft2011}, our bound for \ac{MT-MKL} is tighter, for almost all $M$, when $r$ is small, which is usually a preferred setting.
	
	\item When $r^* \geq \log MT$, the bound given in (\ref{eq:rademacher_learn_feature_r1}) is monotonically increasing with respect to $s$. When $s = \rho^*$, we have $\hat{R}(\mathcal{F}_{\rho^*, r}) \leq \frac{2}{T\sqrt{N}} \sqrt{2eR \log MT}$. This gives a $O(\frac{\sqrt{\log MT}}{T})$ bound. Note that it is proved that the best bound that can be obtained in single-task multiple kernel classification is of order $O(\sqrt{\log M})$ \cite{Cortes2010}. Obviously, this logarithmic bound is preserved in the \ac{MT-MKL} context. When $s$ further decreases to $1$, we have $\hat{R}(\mathcal{F}_{1, r}) \leq \frac{2}{T\sqrt{N}} \sqrt{2R M^{\frac{1}{\log MT}} \log MT }$. Since $M^{\frac{1}{\log MT}}$ can never be larger than $e$, this bound is even tighter than the one obtained, when $s = \rho^*$.
	
	\item When $r^* \leq \log T$, the bound that is given in (\ref{eq:rademacher_learn_feature_r1}) is monotonically increasing with respect to $s$. When $s = \rho^*$, we have $\hat{R}(\mathcal{F}_{\rho^*, r}) \leq \frac{2}{T\sqrt{N}} \sqrt{2eR M^{\frac{1}{r^*}} \log T}$. This gives a $O(\frac{\sqrt{M^{\frac{1}{r^*}}\log T}}{T})$ bound. When $s$ further decreases to $1$,  we have $\hat{R}(\mathcal{F}_{1, r}) \leq \frac{2}{T\sqrt{N}} \sqrt{2R M^{\frac{1}{r^*}} \log T}$. As we can see, it further decreases the bound by a constant factor $e$. 
	
	\item Compared to the optimum bounds that are given in the previous two situations, \ie, $r^* \geq \log MT$ and $r^* \leq \log T$, we can see that, when $r^* \geq \log MT$, we achieve a better bound with respect to $M$, \ie, $O(\sqrt{\log M})$ versus $O(\sqrt{M^{\frac{1}{r^*}}})$. On the other hand, with regards to $T$, even though we get a $O(\frac{\sqrt{\log T}}{T})$ bound in both cases, the case of $r^* \leq \log T$ features a lower constant factor.
	
\end{itemize}

To summarize, \ac{MT-MKL} not only preserves the optimal $O(\sqrt{\log M})$ bound encountered in single-task \ac{MKL}, but also preserves the optimal $O(\frac{\sqrt{\log T}}{T})$ bound encountered in the single-kernel \ac{MTL} case, which was given in the previous section.

\section{Discussion}
\label{sec:Discussion}

\subsection{Relation to Group-Lasso type regularizer}
\label{sec:RelationToGroupLasso}

In the next theorem, we show the relation between our \ac{HS} and the one that is based on Group-Lasso type regularizer.

\begin{theorem}
The \ac{HS} $\mathcal{F}_s$ is equivalent to 

\begin{equation}
\mathcal{F}_{s}^{GL} \triangleq  \{ \boldsymbol{x} \mapsto [ \langle \boldsymbol{w}_1, \phi(\boldsymbol{x}) \rangle, \cdots, \langle \boldsymbol{w}_T, \phi(\boldsymbol{x}) \rangle]' :  (\sum_{t=1}^T \| \boldsymbol{w}_t \|^{s} )^{\frac{2}{s}} \leq R \}
\label{eq:hs_gl_skl}
\end{equation}

%\gcacomment{Ugly: $\mathcal{F}_{s\_gl}$. Nice: $\mathcal{F}_{s}^{\text{GL}}$  :-)}
%\mycomment{Nice :)}

\noindent
Similarly, $\mathcal{F}_{s, r}$ is equivalent to

\begin{equation}
\mathcal{F}_{s,r}^{GL} \triangleq  \{ \boldsymbol{x} \mapsto [ \langle \boldsymbol{w}_1, \tilde{\phi}(\boldsymbol{x}) \rangle, \cdots, \langle \boldsymbol{w}_T, \tilde{\phi}(\boldsymbol{x}) \rangle]' : (\sum_{t=1}^T \| \boldsymbol{w}_t \|^{s} )^{\frac{2}{s}} \leq R, \boldsymbol{\theta} \in \Omega_r(\boldsymbol{\theta}) \}
\label{eq:hs_gl_mkl}
\end{equation}

\noindent
where $\| \boldsymbol{w}_t \|^2 = \sum_{m=1}^M \frac{\| \boldsymbol{w}_t^m \|^2}{\theta_m}$, $\tilde{\phi} = (\phi_1, \cdots, \phi_M)$, $\Omega_r(\boldsymbol{\theta}) = \{ \boldsymbol{\theta} : \boldsymbol{\theta} \succeq \boldsymbol{0}, \| \boldsymbol{\theta} \|_r \leq 1 \}$.

\label{thm:group_lasso}
\end{theorem}

Obviously, by employing the Group-Lasso type regularizer, one can obtain the \ac{HS}s that are proposed in previous sections. Below is a Regularization-Loss framework based on this regularizer with pre-selected kernel:

\begin{equation}
\min_{\boldsymbol{w}_1,\cdots,\boldsymbol{w}_T} \; (\sum_{t=1}^T \| \boldsymbol{w}_t \|^{s} )^{\frac{2}{s}} + C \sum_{i,t} L(\langle \boldsymbol{w}_t, \phi(\boldsymbol{x}_{t}^i) \rangle, y_t^i)
\label{eq:framework_group_lasso}
\end{equation}

\noindent
The \ac{MKL}-based model can be similarly defined.

\subsection{Other related works}
\label{sec:other_works}
There has been substantial efforts put on the research of kernel-based \ac{MTL} and also \ac{MT-MKL}. We in this subsection discuss four closely related papers, and emphasize the difference between this paper and these works.

First, we consider \cite{Aflalo2011} and \cite{Rakotomamonjy2011}. Both of these two papers consider Group-Lasso type regularizer to achieve different level of sparsity. Specifically, \cite{Aflalo2011} utilized the regularizer 

\begin{equation}
(\sum_{j=1}^n (\sum_{k=1}^{n_j} \| \boldsymbol{w}_{jk} \|)^{s})^{\frac{2}{s}}, \; s \geq 2
\label{eq:regularizer_1}
\end{equation}

\noindent
where $l_1$ norm regularization is applied to the weights of each of the $n$ groups (\ie the inner summation), and the group-wise regularization is achieved via $l_s$ norm regularizer. It is hoped that by utilizing this regularizer, one can achieve inner-group sparsity and group-wise non-sparsity. This regularizer can be applied to \ac{MT-MKL}, by letting $\boldsymbol{w}_{jk}$ to be $\boldsymbol{w}_t^m$, which yields the regularizer

\begin{equation}
(\sum_{t=1}^T (\sum_{m=1}^M \| \boldsymbol{w}_t^m \|)^{s})^{\frac{2}{s}}, \; s \geq 2
\label{eq:regularizer_2}
\end{equation}

\noindent
This is similar to the one that appeared in $\mathcal{F}_{s,r}^{GL}$. However, the major difference between our regularizer and (\ref{eq:regularizer_2}) is that, in $\mathcal{F}_{s,r}^{GL}$, instead of applying an $l_1$ norm to the inner summation, we used $\sum_{m=1}^M \frac{\| \boldsymbol{w}_t^m \|^2}{\theta_m}$, where $\theta_m$ has a feasible region that is parametrized by $r$. Therefore, our regularizer encompasses \ac{MT-MKL} with common kernel function, which is learned during training, while (\ref{eq:regularizer_2}) does not. 

In \cite{Rakotomamonjy2011}, the authors considered

\begin{equation}
\sum_{m=1}^M (\sum_{t=1}^T \| \boldsymbol{w}_t^m \|^q)^{\frac{p}{q}}, \; 0 \leq p \leq 1, q \geq 1
\label{eq:regularizer_3}
\end{equation}

\noindent
By applying the $l_p$ (pseudo-)norm, the authors intended to achieve sparsity over the outer summation, while variable sparsity is obtained for the inner summation, due to the $l_q$ norm. The major difference between (\ref{eq:regularizer_3}) and $\mathcal{F}_{s,r}^{GL}$ are twofold. First, the order of the double summation is different, \ie, in (\ref{eq:regularizer_3}), the $\boldsymbol{w}_t^m$'s that belong to the same \ac{RKHS} is considered as a group, while $\mathcal{F}_{s,r}^{GL}$ treats each task as a group. Second, similar to the reason that is discussed above, (\ref{eq:regularizer_3}) does not encompasses the \ac{MT-MKL} with common kernel function, which is learned during the training phase.

In the following, we discuss the difference between our work and the two theoretical works, \cite{Maurer2012} and \cite{Kakade2012}, which derived generalization bound of the \ac{HS}s that are similar to ours. In \cite{Maurer2012}, the authors consider the regularizer

\begin{equation}
\| \boldsymbol{w} \|_{\mathcal{M}} \triangleq \inf \{ \sum_{M \in \mathcal{M}} \| \boldsymbol{v}_M \| : \boldsymbol{v}_M \in \mathcal{H}, \sum_{M \in \mathcal{M}} M \boldsymbol{v}_M = \boldsymbol{w} \}
\label{eq:maurer2012}
\end{equation}

\noindent
where $\mathcal{M}$ is an almost countable set of symmetric bounded linear operators on $\mathcal{H}$. This general form covers several regularizers, such as Lasso, Group-Lasso, weighted Group-Lasso, etc. A key observation is that, in order for a specific regularizer to be covered by this general expression, the regularizer needs to be either summation of several norms, or the infimum of such a summation over a feasible region. For our regularizer $(\sum_{t=1}^T \| \boldsymbol{w}_t \|^{s} )^{\frac{2}{s}}$, obviously, it is not summation of norms (note the power outside the summation). Also, it is not immediately clear, if it can be represented by an infimum, which we just mentioned. Therefore, it appears that there is no succinct way to represent our regularizer as a special case of (\ref{eq:maurer2012}) and the same seems to be the case for our \ac{MT-MKL} regularizer. 

Also, for our \acp{HS}, it is clear how the generalization bounds relate to the number of tasks $T$ and number of kernels $M$, in $\mathcal{F}_s$ and $\mathcal{F}_{s, r}$, and under which circumstances the logarithmic bound can be achieved. This observation may be hard to obtain from the bound that is derived in (\ref{eq:maurer2012}), even though one may view our regularizers as special cases of (\ref{eq:maurer2012}).

In \cite{Kakade2012}, the authors derived generalization bound for regularization-based \ac{MTL} models, with regularizer $\| \boldsymbol{W} \|_{r, p}\triangleq \| (\| \boldsymbol{w}^1\|_r, \cdots, \| \boldsymbol{w}^n\|_r)\|_p$. However, their work assumes $\boldsymbol{W} \in \mathbb{R}^{m \times n}$, while we assume our $\boldsymbol{w}_t$'s be vectors of a potentially infinite-dimensional Hilbert space. Also, such group norm does not generalize our \ac{MT-MKL} regularizer, therefore their bound cannot be applied to our \ac{HS}, even if their results were to be extended to infinite-dimensional vector spaces.

%with a strongly convex regularizer. The authors specifically pointed out that the group norm regularizer

%\gcacomment{Singular or plurals for `bound' and `regularizer'?} 
%\mycomment{I think it should be singular. There is only one bound. But since many models is covered by the HS, so I used plural for ``model''. Since each model has only one regularizer, I used singular for ``regularizer''. Do you think this is correct grammar?} 

%%%%%%%%%%%%%%%%%%%%%%%%%%%%%%%%%%%%%%%%%%%%%%%%%%%%%%%%%%%%%%%%%%%%%%%%%%%%%%%%
%%%%%%%%%%%%%%%%%%%%%%%%%%%%%%%%%%%%%%%%%%%%%%%%%%%%%%%%%%%%%%%%%%%%%%%%%%%%%%%%
%%%%%%%%%%%%%%%%%%%%%%%%%%%%%%%%%%%%%%%%%%%%%%%%%%%%%%%%%%%%%%%%%%%%%%%%%%%%%%%%
\section{Experiments}
\label{sec:Experiments}
In this section, we investigate via experimentation the generalization bounds of our \acp{HS}. We first evaluate the discrepancy between the \ac{ERC} of $\mathcal{F}_s$, $\mathcal{F}_{s, r}$ and their bounds. We show experimentally that the bound gives a good estimate of the relevant \ac{ERC}. Then, we consider a new \ac{SVM}-based \ac{MTL} model that uses $\mathcal{F}_{s}$ as its \ac{HS}. The model is subsequently extended to allow for \ac{MT-MKL} by using $\mathcal{F}_{s, r}$ as its \ac{HS}. 

%The latter model will be evaluated on four different multi-task data sets.

%%%%%%%%%%%%%%%%%%%%%%%%%%%%%%%%%%%%%%%%%%%%%%%%%%%%%%%%%%%%%%%%%%%%%%%%%
%%%%%%%%%%%%%%%%%%%%%%%%%%%%%%%%%%%%%%%%%%%%%%%%%%%%%%%%%%%%%%%%%%%%%%%%%
\subsection{\ac{ERC} Bound Evaluation}
\label{sec:Evaluate_bound}

%In this subsection, we numerically evaluate the difference between our bound and the real \ac{ERC} for the two \ac{HS} $\mathcal{F}_s$ and $\mathcal{F}_{s, r}$. 
For $\mathcal{F}_s$, given a data set and a pre-selected kernel function, we can calculate its kernel matrices $\boldsymbol{K}_t, t=1,\cdots,T$. Then, the \ac{ERC} is given by \eref{eq:rademacher_alternative}. In order to approximate the expectation $E_\sigma \{\left \| \boldsymbol{u} \right \|_{s^*}\}$, we resort to Monte Carlo simulation by drawing a large number, $D$, of i.i.d samples for the $\boldsymbol{\sigma}_t$'s from a uniform distribution on the hyper-cube $\left\{ -1, 1\right\}^N$. Subsequently, for each sample we evaluate the argument of the expectation and average the results. For $\mathcal{F}_{s, r}$, the \ac{ERC} is calculated as in the first equation of (\ref{eq:rademacher_alternative_learn_feature_1}). For each of the $D$ samples of $\boldsymbol{\sigma}_t$, we can calculate the corresponding $\boldsymbol{u}_t$. Then, we solve the maximization problem by using CVX \cite{Grant2008,cvx}. Finally, we calculate the average of the $D$ values to approximate the \ac{ERC}. For the experiment related to $\mathcal{F}_{s, r}$, we only considered the case, when $s \geq 2$. Under these circumstances, 
the maximization problem in (\ref{eq:rademacher_alternative_learn_feature_1}) is concave and can be easily solved, unlike the case, when $s \in [1, 2)$.

We used the Letter data set \footnote{Available at: \url{http://www.cis.upenn.edu/~taskar/ocr/}} for this set of experiments. It is a collection of handwritten words compiled by Rob Kassel of the MIT Spoken Language Systems Group. The associated \ac{MTL} problem involves $8$ tasks, each of which is a binary classification problem for handwritten letters. The $8$ tasks are: `C' \vs\ `E', `G' \vs\ `Y', `M' \vs\ `N', `A' \vs\ `G', `I' \vs\ `J', `A' \vs\ `O', `F' \vs\ `T' and `H' \vs\ `N'. Each letter is represented by a $8 \times 16$ pixel image, which forms a $128$-dimensional feature vector. We chose $100$ samples for each letter, and set $D = 10^4$. To calculate the kernel matrix, we used a Gaussian kernel with spread parameter $2^7$ for $\mathcal{F}_s$, and 9 different Gaussian kernels with spreads $\{ 2^{-7}, 2^{-5}, 2^{-3}, 2^{-1}, 2^{0}, 2^{1}, 2^{3}, 2^{5}, 2^{7} \}$ for $\mathcal{F}_{s, r}$. Finally, $R$ was set to $1$.

%\begin{figure}[ht]
%\centering
	%\subfloat[HS: $\mathcal{F}_s$]{
		%\includegraphics[width=6cm]{bound_compare_letter_SKL}
		%\label{fig:fs}}
	%\subfloat[HS: $\mathcal{F}_{s, r}$]{
		%\includegraphics[width=6cm]{bound_compare_letter_MKL}
		%\label{fig:fsr}}
%\caption{Comparison between Monte Carlo-estimated \acp{ERC} and our derived bounds using the Letter data set. $10^4$ $\boldsymbol{\sigma}_t$ samples were used for Monte Carlo estimation. We sampled $100$ data for each letter and used $9$ kernel functions in multiple kernel scenario.}
%\label{fig:discrepancy_erc_bound}
%\end{figure}

\begin{figure}[ht]
\centering
	\subfloat[HS: $\mathcal{F}_s$]{
		\includegraphics[width=8cm]{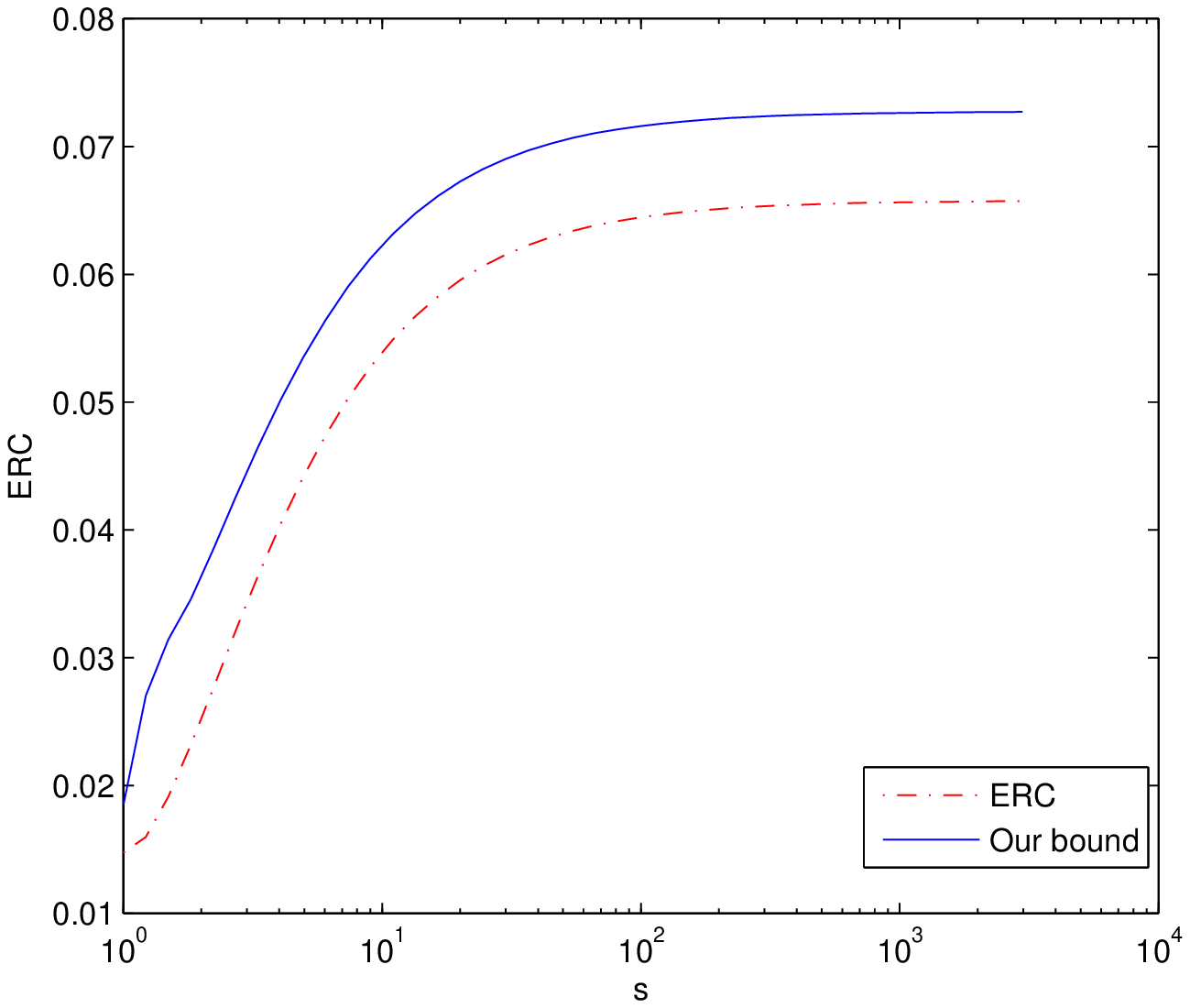}
		\label{fig:fs}}
	\subfloat[HS: $\mathcal{F}_{s, r}$]{
		\includegraphics[width=8cm]{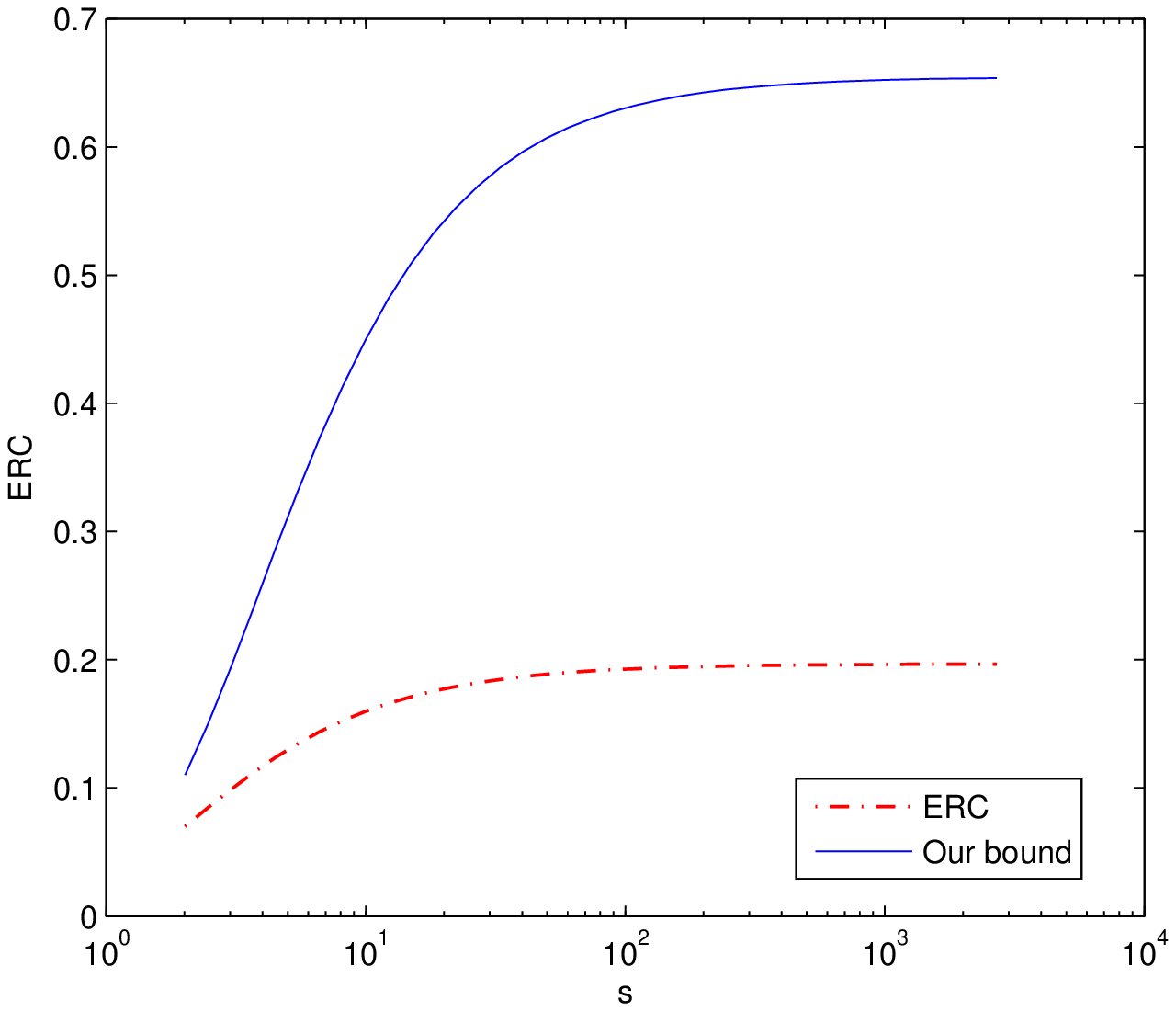}
		\label{fig:fsr}}
\caption{Comparison between Monte Carlo-estimated \acp{ERC} and our derived bounds using the Letter data set. $10^4$ $\boldsymbol{\sigma}_t$ samples were used for Monte Carlo estimation. We sampled $100$ data for each letter and used $9$ kernel functions in multiple kernel scenario.}
\label{fig:discrepancy_erc_bound}
\end{figure}

%\gcacomment{Provide more information in the figure caption on what settings you used to obtain these plots.}
%\mycomment{See how it looks now.}

The experimental results are shown in \fref{fig:discrepancy_erc_bound}. In both sub-figures, it is obvious that both our bound and the real \ac{ERC} are monotonically increasing. For $\mathcal{F}_s$, it can be seen that the bound is tight everywhere. For $\mathcal{F}_{s, r}$, even though the difference between our bound and the Monte Carlo estimated \ac{ERC} becomes larger when $s$ grows, the bound is still tight for small $s$. %Due to the theoretical analysis that is given before, as we can expect, the tight bound can also be obtained when $s \in [1, 2)$. 
This experiment shows a good match between the real \ac{ERC} and our bound, which verifies our theoretical analysis in \sref{sec:FixedFeatureMapping} and \sref{sec:LearningFeatureMapping}.%For any other data sets, similar results can be expected.

%%%%%%%%%%%%%%%%%%%%%%%%%%%%%%%%%%%%%%%%%%%%%%%%%%%%%%%%%%%%%%%%%%%%%%%%%
%%%%%%%%%%%%%%%%%%%%%%%%%%%%%%%%%%%%%%%%%%%%%%%%%%%%%%%%%%%%%%%%%%%%%%%%%
\subsection{\ac{SVM}-based Model}
\label{sec:Model}

In this subsection we present a new \ac{SVM}-based model which reflects our proposed \ac{HS}. For training data $\{\boldsymbol{x}_t^i, y_t^i \} \in \mathcal{X} \times \{-1, 1\}, i = 1,\cdots,N_t, t = 1,\cdots,T$ and fixed feature mapping $\phi : \mathcal{X} \mapsto \mathcal{H}$, our model is given as follows:

\begin{equation}
\begin{aligned}
\min_{\boldsymbol{w}, \boldsymbol{\xi}, b}& (\sum_{t=1}^T (\frac{\| \boldsymbol{w}_t \|^2}{2})^{\frac{s}{2}} )^{\frac{2}{s}} + C \sum_{t, i= 1} ^ {T, N_t} \xi_t^i\\
\text{s.t.} & y_t^i (\left \langle \boldsymbol{w}_t, \phi(\boldsymbol{x}_t^i) \right \rangle + b_t) \geq 1-\xi_t^i , \;\;\xi_t^i \geq 0, \forall i, t 
\end{aligned}
\label{eq:model_fix_feature}
\end{equation}

\noindent
Obviously, $\mathcal{F}_s$ is the \ac{HS} of (\ref{eq:model_fix_feature}). Such minimization problem can be solved as follows. First, note that when $1 \leq s \leq 2$, the problem is equivalent to 

\begin{equation}
\begin{aligned}
\min_{\boldsymbol{w}, \boldsymbol{\xi}, \boldsymbol{b}, \boldsymbol{\lambda}} & \sum_{t=1}^T \frac{\| \boldsymbol{w}_t \|^2}{2 \lambda_t} + C\sum_{t, i= 1} ^ {T, N_t} \xi_t^i\\
\text{s.t.} & y_t^i (\left \langle \boldsymbol{w}_t, \phi(\boldsymbol{x}_t^i) \right \rangle + b_t) \geq 1-\xi_t^i , \;\;\xi_t^i \geq 0, \forall i, t \\
 & \boldsymbol{\lambda} \succeq \boldsymbol{0}, \left \| \boldsymbol{\lambda} \right \|_{\frac{s}{2-s}} \leq 1
\end{aligned}
\label{eq:model_fix_feature_equi_1}
\end{equation}

\noindent
which can be easily solved via block coordinate descent method, with $\{ \boldsymbol{w}, \boldsymbol{\xi}, \boldsymbol{b} \}$ as a group and $\boldsymbol{\lambda}$ as another. When $s > 2$, (\ref{eq:model_fix_feature}) is equivalent to

\begin{equation}
\begin{aligned}
\min_{\boldsymbol{w}, \boldsymbol{\xi}, \boldsymbol{b}} \max_{\boldsymbol{\lambda}} & \sum_{t=1}^T \frac{\lambda_t \| \boldsymbol{w}_t \|^2}{2} + C\sum_{t, i= 1} ^ {T, N_t} \xi_t^i\\
\text{s.t.} & y_t^i (\left \langle \boldsymbol{w}_t, \phi(\boldsymbol{x}_t^i) \right \rangle + b_t) \geq 1-\xi_t^i , \;\;\xi_t^i \geq 0, \forall i, t \\
 & \boldsymbol{\lambda} \succeq \boldsymbol{0}, \left \| \boldsymbol{\lambda} \right \|_{\frac{s}{s-2}} \leq 1
\end{aligned}
\label{eq:model_fix_feature_equi_2}
\end{equation}

\noindent
Since it is a convex-concave min-max problem with compact feasible region, the order of min and max can be interchanged \cite{Sion1958}, which gives the objective function

\begin{equation}
\max_{\boldsymbol{\lambda}} \min_{\boldsymbol{w}, \boldsymbol{\xi}, \boldsymbol{b}} \sum_{t=1}^T \lambda_t ( \frac{\| \boldsymbol{w}_t \|^2}{2} + \frac{C}{\lambda_t} \sum_{i=1}^{N_t} \xi_t^i)
\label{eq:model_fix_feature_equi_3}
\end{equation}

\noindent 
Calculating the dual form of the inner \ac{SVM} problem gives the following maximization problem:

\begin{equation}
\begin{aligned}
\max_{\boldsymbol{\alpha}, \boldsymbol{\lambda}}\; & \sum_{t=1}^T \lambda_t (\boldsymbol{\alpha}_t' \boldsymbol{1} - \frac{1}{2} \boldsymbol{\alpha}_t' \boldsymbol{Y}_t \boldsymbol{K}_t \boldsymbol{Y}_t \boldsymbol{\alpha}_t) \\
\text{s.t.} & \boldsymbol{0} \preceq \boldsymbol{\alpha}_t \preceq \frac{C}{\lambda_t}\boldsymbol{1},\; \boldsymbol{\alpha}_t' \boldsymbol{y}_t = 0,\; \forall t \\
 & \boldsymbol{\lambda} \succeq \boldsymbol{0}, \left \| \boldsymbol{\lambda} \right \|_{\frac{s}{s-2}} \leq 1
\end{aligned}
\label{eq:model_fix_feature_equi_4}
\end{equation}

\noindent
where $\boldsymbol{Y}_t \triangleq \textit{diag}([y_t^1, \cdots, y_t^{N_t}]')$ and $\boldsymbol{K}_t$ is the kernel matrix that is calculated based on the training data from the $t$-th task. Group coordinate descent can be utilized to solve (\ref{eq:model_fix_feature_equi_4}), with $\boldsymbol{\lambda}$ as a group and $\boldsymbol{\alpha}$ as another group. 
%In our experiments, for fixed $\boldsymbol{\alpha}$, we optimize (\ref{eq:model_fix_feature_equi_4}) with respect to $\boldsymbol{\lambda}$ by utilizing CVX \cite{cvx} \cite{Grant2008}. For fixed $\boldsymbol{\lambda}$, $T$ \ac{SVM} problems are involved, and in our experiments, they are solve by LIBSVM \cite{CC01a}.

The model can be extended so that it can accommodate \ac{MKL} as follows:

\begin{equation}
\begin{aligned}
\min_{\boldsymbol{w}_t, \boldsymbol{\xi}_t, b_t, \boldsymbol{\theta}}& (\sum_{t=1}^T (\sum_{m=1}^M \frac{\| \boldsymbol{w}_t^m \|^2}{2 \theta_m})^{\frac{s}{2}} )^{\frac{2}{s}} + C\sum_{t, i= 1} ^ {T, N_t} \xi_t^i\\
\text{s.t.} & y_t^i (\left \langle \boldsymbol{w}_t, \phi(\boldsymbol{x}_t^i) \right \rangle + b_t) \geq 1-\xi_t^i , \;\;\xi_t^i \geq 0, \forall i, t \\
 & \boldsymbol{\theta} \succeq \boldsymbol{0},  \left \| \boldsymbol{\theta} \right \|_r \leq 1
\end{aligned}
\label{eq:model_learn_feature}
\end{equation}

\noindent
where $\phi =  (\phi_1, \cdots, \phi_M)$. Obviously, its \ac{HS} is $\mathcal{F}_{s, r}$. This model can be solved via the similar strategy of solving (\ref{eq:model_fix_feature}). The only situation that needs a different algorithm is the case when $s > 2$, where (\ref{eq:model_learn_feature}) will be transformed to

\begin{equation}
\begin{aligned}
\min_{\boldsymbol{\theta}} \max_{\boldsymbol{\alpha}, \boldsymbol{\lambda}}\; & \sum_{t=1}^T \lambda_t (\boldsymbol{\alpha}_t' \boldsymbol{1} - \frac{1}{2} \boldsymbol{\alpha}_t' \boldsymbol{Y}_t (\sum_{m=1}^M \theta_m \boldsymbol{K}_t^m) \boldsymbol{Y}_t \boldsymbol{\alpha}_t)  \\
\text{s.t.} & \boldsymbol{0} \preceq \boldsymbol{\alpha}_t \preceq \frac{C}{\lambda_t}\boldsymbol{1},\; \boldsymbol{\alpha}_t' \boldsymbol{y}_t = 0,\; \forall t \\
 & \boldsymbol{\lambda} \succeq \boldsymbol{0}, \left \| \boldsymbol{\lambda} \right \|_{\frac{s}{s-2}} \leq 1 \\
 & \boldsymbol{\theta} \succeq \boldsymbol{0}, \left \| \boldsymbol{\theta} \right \|_r \leq 1 
\end{aligned}
\label{eq:model_learn_feature_equi_1}
\end{equation}

\noindent
This min-max problem cannot be solved via group coordinate descent. Instead, we use the Exact Penalty Function method to solve it. We omit the details of this method and refer the readers to \cite{Watson1981}, since it is not the focus of this paper.

%%%%%%%%%%%%%%%%%%%%%%%%%%%%%%%%%%%%%%%%%%%%%%%%%%%%%%%%%%%%%%%%%%%%%%%%%
%%%%%%%%%%%%%%%%%%%%%%%%%%%%%%%%%%%%%%%%%%%%%%%%%%%%%%%%%%%%%%%%%%%%%%%%%
\subsection{Experimental Results on the \ac{SVM}-based model}
\label{sec:results_SVM}

We performed our experiments on two well-known and frequently-used multi-task data sets, namely Letter and Landmine, and two handwritten digit data sets, namely MNIST and USPS. The Letter data set was described in the previous sub-section. Due to the large size of the original Letter data set, we randomly sampled $200$ points for each letter to construct a training set. One exception is the letter $j$, as it contains only $189$ samples in total. The Landmine data set\footnote{Available at: \url{http://people.ee.duke.edu/~lcarin/LandmineData.zip}} consists of $29$ binary classification tasks. Each datum is a $9$-dimensional feature vector extracted from radar images that capture a single region of landmine fields. Tasks $1-15$ correspond to regions that are relatively highly foliated, while the other $14$ tasks correspond to regions that are bare earth or desert. The tasks entail different amounts of data, varying from $30$ to $96$ samples. The goal is to detect landmines in specific regions. 

Regarding the MNIST \footnote{Available at: \url{http://yann.lecun.com/exdb/mnist/}} and USPS \footnote{Available at: \url{http://www.cs.nyu.edu/~roweis/data.html}} data sets, each of the two are grayscale images containing handwritten digits from $0$ to $9$ with $784$ and $256$ features respectively. As was the case with the Letter data set, due to the large size of the original data set, we randomly sampled $100$ data from each digit population to form a training set consisting of $1000$ samples in total. To simulate the \ac{MTL} scenario, we split the data into $45$ binary classification tasks by applying a one-versus-one strategy. The classification accuracy was then calculated as the average of classification accuracies over all tasks. 

\begin{figure}[ht]
\centering
	\subfloat[Fixed kernel (single kernel) scenario]{
		\includegraphics[width=8cm]{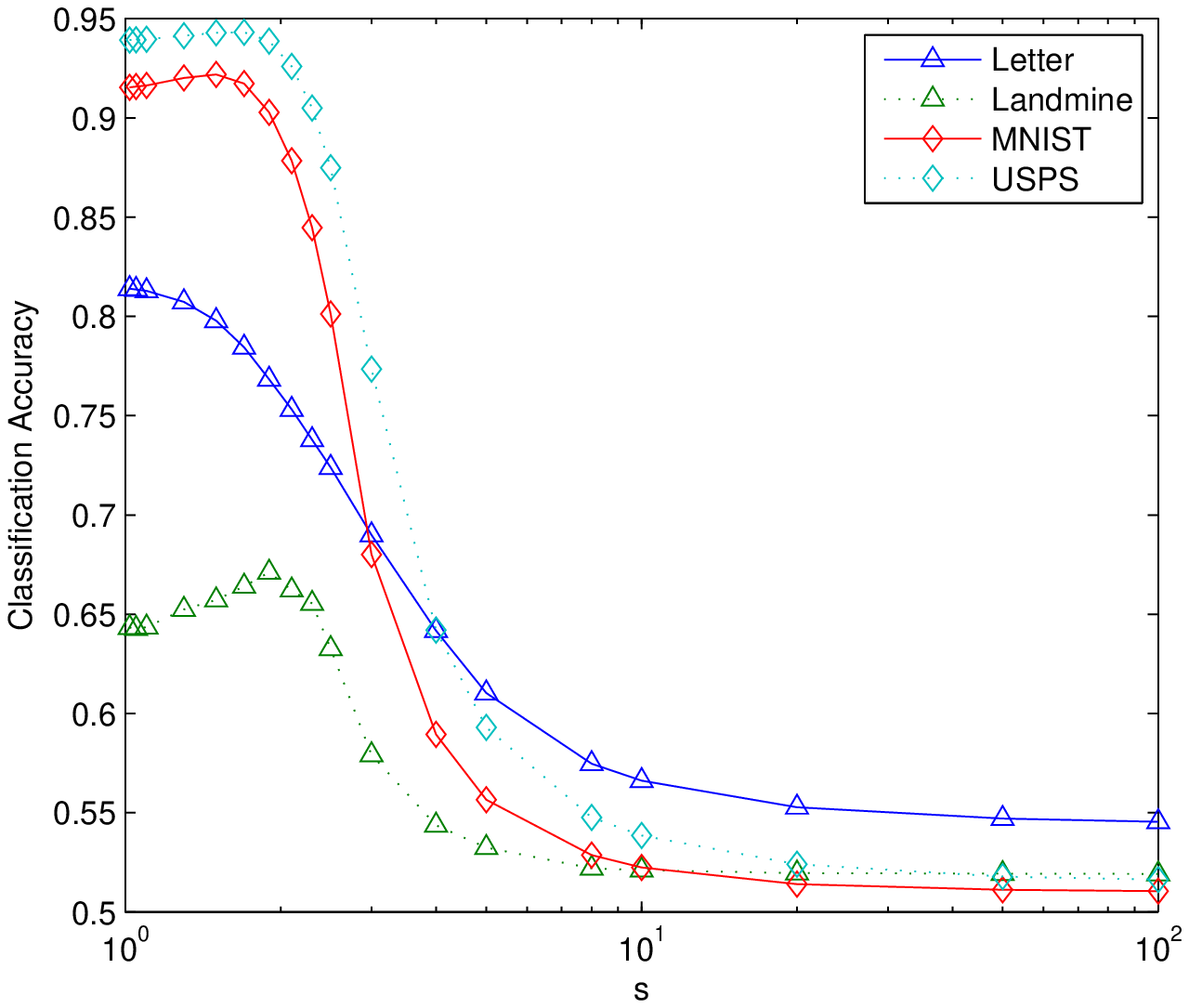}
		\label{fig:skl}}
	\subfloat[Multiple kernel learning scenario]{
		\includegraphics[width=8cm]{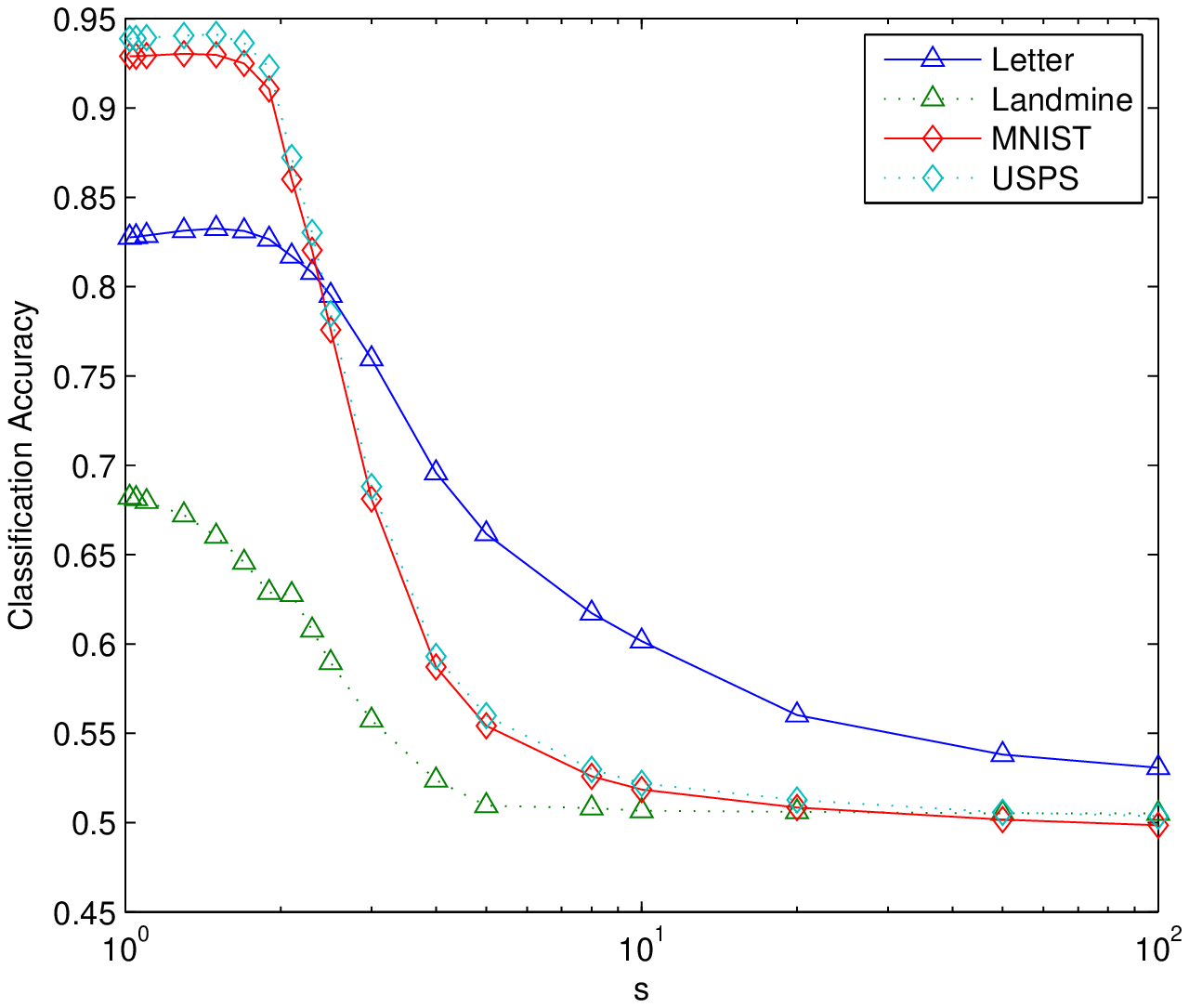}
		\label{fig:mkl}}
\caption{Average classification accuracy on $20$ runs for different $s$ values}
\label{fig:svm_results}
\end{figure}

For all our experiments, the training set size was set to $10\%$ of the available data. We did not choose large training sets, since, as we can see from the generalization bound in (\ref{eq:rademacher_fix_feature}), (\ref{eq:rademacher_learn_feature}) and (\ref{eq:rademacher_learn_feature_r1}), when $N$ is large, the effect of $s$ becomes minor. For \ac{MT-MKL}, we chose the $9$ Gaussian kernels that were introduced in \sref{sec:Evaluate_bound}, as well as a linear and a $2^{nd}$-order polynomial kernel. For the single kernel case, we selected the optimal kernel from these 11 kernel function candidates via cross-validation. \ac{SVM}'s regularization parameter $C$ was selected from the set $\{1/81, 1/27, 1/9, 1/3, 1, 3, 9, 27, 81\}$. In the \ac{MT-MKL} case, the norm parameter $r$ for $\boldsymbol{\theta}$ was set to $1$ to induce sparsity on $\boldsymbol{\theta}$. We varied $s$ from $1$ to $100$, and reported the best average classification accuracy over $20$ runs. For $s > 100$, the results are almost always the same as that when $s = 100$, therefore we did not report these results. In fact, as show below, the model performance deteriorates quickly when $s > 2$, and changes very few when $s > 10$. The experimental results are given in \fref{fig:svm_results}.

%, when $s \leq 10000$, and compared it to the accuracy obtained, when $s = 10000$. We used the value of $s = 10000$ as an approximation to $s = +\infty$. We also reported $s_{opt}$, the $s$ value that gives the highest classification accuracy. The results are given in \tref{Table:results}. 

%\begin{table}[ht]
%\begin{center}
%\caption{Average classification accuracies on the four data sets} 
%\label{Table:results}
%\begin{tabular}{lcccc}
%\toprule 
%$\mathcal{F}_s$ & Letter & Landmine & MNIST & USPS  \\
%\midrule
%$s \leq +\infty$ & 86.36 & 68.21 & 92.69 & 94.21\\
%$s = +\infty$ & 83.85& 68.21 & 92.69& 94.21 \\
%$s_{opt}$ & 1 & $+\infty$ & $+\infty$ & $+\infty$ \\
%\midrule
%$\mathcal{F}_{s, r}$ & Letter & Landmine & MNIST & USPS  \\
%\midrule
%$s \leq +\infty$ & 83.05 & 64.34 & 93.08 & 93.26 \\
%$s = +\infty$ & 80.53 & 62.57 & 90.69 & 90.59 \\
%$s_{opt}$ & 1 & 1 & 1 & 1 \\
%\bottomrule
%\end{tabular}
%\end{center}
%\end{table}

It can be seen that, the classification accuracy is roughly monotonically decreasing with respect to $s$, and the performance deteriorates significantly when $s > 2$. In many situations, the best performance is achieved, when $s = 1$. This result supports our theoretical analysis that the lowest generalization bound is obtained when $s = 1$. On the other hand, in some situations, such as the cases which consider the USPS data set in a single kernel setting, and the Letter data set in multiple kernel setting, the optimum model is not obtained when $s = 1$. This seems contradictory to our previously stated claims. However, this phenomenon can be explained similarly to the discussion in Section $5.1$ of \cite{Kloft2011}, which we summarize here: Obviously, for different $s$, the optimal solution ($f_t$'s) may be different. To get the optimal solution, we need to tune the ``size'' of the \ac{HS}, such that the optimum $f_t$'s are contained in the \ac{HS}. This implies that the size of the \ac{HS}, which is parametrized by $R$, could be different for different $s$, instead of being fixed as discussed in previous sections. It is possible that the \ac{HS} size (thus $R$) is very small, when $s \neq 1$. In this scenario, the lowest bound could be obtained when $s \neq 1$. For a more detailed discussion, we refer the reader to Section $5.1$ in \cite{Kloft2011}. Finally, it is interesting to see how the performance deteriorates when $s$ becomes large. The reason for the bad performance is as follows. Observe that the regularizer is the $l_{\frac{s}{2}}$ norm of the $T$ \ac{SVM} regularizers. Consider the extreme case, when $s \rightarrow \infty$, the $l_{\frac{s}{2}}$ norm becomes the $l_{\infty}$ norm, which is $\max_{t} \{ \frac{\| \boldsymbol{w}_t \|^2}{2} \}$. In this scenario, the regularizer of the model is only the one which has the smallest margin, while the regularizers of other tasks are ignored. Therefore, it is not a surprise that the performance of the other tasks is bad, which leads to low average classification accuracy. For large $s$ value, even though it is not infinity, the bad result can be similarly analyzed.

%%%%%%%%%%%%%%%%%%%%%%%%%%%%%%%%%%%%%%%%%%%%%%%%%%%%%%%%%%%%%%%%%%%%%%%%%%%%%%%%
%%%%%%%%%%%%%%%%%%%%%%%%%%%%%%%%%%%%%%%%%%%%%%%%%%%%%%%%%%%%%%%%%%%%%%%%%%%%%%%%
%%%%%%%%%%%%%%%%%%%%%%%%%%%%%%%%%%%%%%%%%%%%%%%%%%%%%%%%%%%%%%%%%%%%%%%%%%%%%%%%

% Reset all acronyms
\acresetall

\section{Conclusions}
\label{sec:Conclusions}

%\gcacomment{Did not touch; please see my comments, expressions I use, etc. and revise it wherever necessary... (e.g. ``the O() bound''}
%\mycomment{Did some changes.}
%\gcacomment{I have revised it; please check it...}

In this paper, we proposed a \ac{MTL} \ac{HS} $\mathcal{F}_s$ involving $T$ discriminative functions parametrized by weights $\boldsymbol{w}_t$. The weights are controlled by norm-ball constraints, whose radii are variable and estimated during the training phase. It extends an \ac{HS} $\tilde{\mathcal{F}}$ that has been previously investigated in the literature, where the radii are pre-determined. It is shown that the latter space is a special case of $\mathcal{F}_s$, when $s \rightarrow +\infty$. We derived and analyzed the generalization bound of $\mathcal{F}_s$ and have shown that the bound is monotonically increasing with respect to $s$. Also, in the optimal case ($s=1$), a bound of order $O(\frac{\sqrt{\log T}}{T})$ is achieved. We further extended the \ac{HS} to $\mathcal{F}_{s, r}$, which is suitable for \ac{MT-MKL}. Similar results were obtained, including a bound that is monotonically increasing with $s$ and an optimal bound of order $O(\frac{\sqrt{\log MT}}{T})$, when $s=1$. The experimental results shown that our \ac{ERC} bound is tight and matches the real \ac{ERC} very well. We then demonstrated the relation between our \ac{HS} and the Group-Lasso type regularizer, and a \ac{SVM}-based model was proposed with \ac{HS} $\mathcal{F}_s$, that was further extended to handle \ac{MT-MKL} by using the \ac{HS} $\mathcal{F}_{s, r}$. The experimental results on multi-task classification data sets showed that the classification accuracy is monotonically decreasing with respect to $s$, and the optimal results for most experiments are indeed achieved, when $s = 1$, as indicated by our analysis. The presence of results that, contrary to our analysis, are optimal, when $s \neq 1$, can be justified similarly to Section $5.1$ in \cite{Kloft2011}.

\section*{Acknowledgments}
\addcontentsline{toc}{section}{Acknowledgments}

C. Li acknowledges partial support from \ac{NSF} grant No. 0806931. Moreover, M. Georgiopoulos acknowledges partial support from \ac{NSF} grants No. 0525429, No. 0963146, No. 1200566 and No. 1161228. Finally, G. C. Anagnostopoulos acknowledges partial support from \ac{NSF} grants No. 0717674 and No. 0647018. Any opinions, findings, and conclusions or recommendations expressed in this material are those of the authors and do not necessarily reflect the views of the \ac{NSF}.

% The plainrul style will printout the URL field of each bib entry
% and hyperref will create a clickable link.
\bibliographystyle{plainurl}
% Modify the bibliography file name as necessary.
\bibliography{ArXiv2013paperB}

\begin{thebibliography}{10}

\bibitem{Aflalo2011}
Jonathan Aflalo, Aharon Ben-Tal, Chiranjib Bhattacharyya, Jagarlapudi~Saketha
  Nath, and Sankaran Raman.
\newblock Variable sparsity kernel learning.
\newblock {\em Journal of Machine Learning Research}, 12:565--592, 2011.

\bibitem{Argyriou2008}
Andreas Argyriou, Theodoros Evgeniou, and Massimiliano Pontil.
\newblock Convex multi-task feature learning.
\newblock {\em Machine Learning}, 73:243--272, 2008.

\bibitem{Caponnetto2008}
Andrea Caponnetto, Charles~A. Micchelli, Massimiliano Pontil, and Yiming Ying.
\newblock Universal multi-task kernels.
\newblock {\em Journal of Machine Learning Research}, 9:1615--1646, 2008.

\bibitem{Caruana1997}
Rich Caruana.
\newblock Multitask learning.
\newblock {\em Machine Learning}, 28:41--75, 1997.

\bibitem{Chen2011}
Jianhui Chen, Jiayu Zhou, and Jieping Ye.
\newblock Integrating low-rank and group-sparse structures for robust
  multi-task learning.
\newblock In {\em KDD}, 2011.

\bibitem{Cortes2010}
Corinna Cortes, Mehryar Mohri, and Afshin Rostamizadeh.
\newblock Generalization bounds for learning kernels.
\newblock In {\em ICML}, 2010.

\bibitem{Evgeniou2005}
Theodoros Evgeniou, Charles~A. Micchelli, and Massimiliano Pontil.
\newblock Learning multiple tasks with kernel methods.
\newblock {\em Journal of Machine Learning Research}, 6:615--637, 2005.

\bibitem{Fei2011}
Hongliang Fei and Jun Huan.
\newblock Structured feature selection and task relationship inference for
  multi-task learning.
\newblock In {\em ICDM}, 2011.

\bibitem{Gonen2011}
Mehmet Gonen and Ethem Alpaydin.
\newblock Multiple kernel learning algorithms.
\newblock {\em Journal of Machine Learning Research}, 12:2211--2268, 2011.

\bibitem{Gong2012}
Pinghua Gong, Jieping Ye, and Changshui Zhang.
\newblock Multi-stage multi-task feature learning.
\newblock In {\em NIPS}, 2012.

\bibitem{Grant2008}
M.~Grant and S.~Boyd.
\newblock Graph implementations for nonsmooth convex programs.
\newblock In V.~Blondel, S.~Boyd, and H.~Kimura, editors, {\em Recent Advances
  in Learning and Control}, Lecture Notes in Control and Information Sciences,
  pages 95--110. Springer-Verlag Limited, 2008.

\bibitem{cvx}
M.~Grant and S.~Boyd.
\newblock {CVX}: Matlab software for disciplined convex programming, version
  1.21, April 2011.

\bibitem{Kakade2012}
Sham~M. Kakade, Shai Shalev-Shwartz, and Ambuj Tewari.
\newblock Regularization techniques for learning with matrices.
\newblock {\em Journal of Machine Learning Research}, 13:1865--1890, 2012.

\bibitem{Kang2011}
Zhuoliang Kang, Kristen Grauman, and Fei Sha.
\newblock Learning with whom to share in multi-task feature learning.
\newblock In {\em ICML}, 2011.

\bibitem{Kloft2011}
Marius Kloft, Ulf Brefeld, Soren Sonnenburg, and Alexander Zien.
\newblock $l_p$-norm multiple kernel learning.
\newblock {\em Journal of Machine Learning Research}, 12:953--997, 2011.

\bibitem{Kolar2012}
Mladen Kolar and Han Liu.
\newblock Marginal regression for multitask learning.
\newblock In {\em NIPS}, 2012.

\bibitem{Kwapien1992}
Stanislaw Kwapien and Wojbor~A. Woyczynski.
\newblock {\em Random Series and Stochastic Integrals: Single and Multiple}.
\newblock Birkhauser, 1992.

\bibitem{Lanckriet2004}
Gert R.~G. Lanckriet, Nello Cristianini, Peter Bartlett, Laurent~El Ghaoui, and
  Michael~I. Jordan.
\newblock Learning the kernel matrix with semidefinite programming.
\newblock {\em J. Mach. Learn. Res.}, 5:27--72, December 2004.

\bibitem{Lozano2012}
Aurelie~C. Lozano and Grzegorz Swirszcz.
\newblock Multi-level lasso for sparse multi-task regression.
\newblock In {\em NIPS}, 2012.

\bibitem{Maurer2006a}
Andreas Maurer.
\newblock Bounds for linear multi-task learning.
\newblock {\em Journal of Machine Learning Research}, 7:117--139, 2006.

\bibitem{Maurer2006}
Andreas Maurer.
\newblock The rademacher complexity of linear transformation classes.
\newblock In Gábor Lugosi and HansUlrich Simon, editors, {\em Learning Theory},
  volume 4005 of {\em Lecture Notes in Computer Science}, pages 65--78.
  Springer Berlin Heidelberg, 2006.
\newblock \href {http://dx.doi.org/10.1007/11776420_8}
  {\path{doi:10.1007/11776420_8}}.

\bibitem{Maurer2012}
Andreas Maurer and Massimiliano Pontil.
\newblock Structured sparsity and generalization.
\newblock {\em Journal of Machine Learning Research}, 13:671--690, 2012.

\bibitem{Parameswaran2012}
Shibin Parameswaran and Kilian~Q. Weinberger.
\newblock Large margin multi-task metric learning.
\newblock In {\em NIPS}, 2012.

\bibitem{Rakotomamonjy2011}
Alain Rakotomamonjy, Remi Flamary, Gilles Gasso, and Stephane Canu.
\newblock $l_p-l_q$ penalty for sparse linear and sparse multiple kernel
  multitask learning.
\newblock {\em IEEE Transactions on Neural Networks}, 22:1307--1320, 2011.

\bibitem{Sion1958}
M.~Sion.
\newblock On general minimax theorems.
\newblock {\em Pacific Journal of Mathematics}, 8:171--176, 1958.

\bibitem{Tang2009}
Lei Tang, Jianhui Chen, and Jieping Ye.
\newblock On multiple kernel learning with multiple labels.
\newblock In {\em IJCAI}, 2009.

\bibitem{Watson1981}
G.A. Watson.
\newblock Globally convergent methods for semi-infinite programming.
\newblock {\em BIT Numerical Mathematics}, 21:392--373, 1981.

\bibitem{Zhang2010}
Yu~Zhang and Dit-Yan Yeung.
\newblock Transfer metric learning by learning task relationships.
\newblock In {\em KDD}, 2010.

\bibitem{Zhong2012}
Leon~wenliang Zhong and James~T. Kwok.
\newblock Convex multitask learning with flexible task clusters.
\newblock In {\em ICML}, 2012.

\end{thebibliography}

% Remove if you don't have appendix.
\section{Appendix}
\label{sec:appendix}
%%%%%%%%%%%%%%%%%%%%%%%%%%%%%%%%%%%%%%%%%%%%%%
%%%%%%%%%%%%%%%%%%%%%%%%%%%%%%%%%%%%%%%%%%%%%%
\subsection{Preliminaries}
\label{sec:preliminaries_appendix}

In this subsection, we provide two results that will be used in the following subsections.

\begin{lemma}
Let $p \geq 1$, $\boldsymbol{x}, \boldsymbol{a} \in \mathbb{R}^n$ such that $\boldsymbol{a} \succeq \boldsymbol{0}$ and $\boldsymbol{a} \neq \boldsymbol{0}$. Then, 

\begin{equation}
	\max_{\boldsymbol{x} \in \Omega(\boldsymbol{x})} \boldsymbol{a}' \boldsymbol{x} = \left \| \boldsymbol{a} \right \|_p
	\label{eq:ACOne}
\end{equation} 

\noindent
where $\Omega(\boldsymbol{x}) \triangleq \{ \boldsymbol{x} : \| \boldsymbol{x}\|_{p^*} \leq 1\}$.
\label{lemma:simple_max_problem}
\end{lemma}

This lemma can be simply proved by utilizing Lagrangian multiplier method with respect to the maximization problem.

\begin{lemma}
Let $\boldsymbol{x}_1, \cdots, \boldsymbol{x}_n \in \mathcal{H}$, then we have that

\begin{equation}
E_{\sigma} \| \sum_{i=1}^n \sigma_i \boldsymbol{x}_i \|^p \leq  (p \sum_{i=1}^n \| \boldsymbol{x}_i \|^2)^{\frac{p}{2}}
\end{equation}

\noindent
for any $p \geq 1$, where $\sigma_i$'s are the Rademacher-distributed random variables.
\label{lemma:k-k}
\end{lemma}

\noindent
For $1 \leq p < 2$, the above result can be simply proved using Lyapunov's inequality. When $p \geq 2$, the lemma can be proved by using Proposition $3.3.1$ and $3.4.1$ in \cite{Kwapien1992}. %It is worth mentioning that \cite{Kloft2012} did not cite the above lemma correctly.

\subsection{Proof to \lemmaref{lemma:rademacher_alternative_fix_feature}}
\label{sec:Proof_to_rademacher_alternative_fix_feature}
\begin{proof}
First notice that $\mathcal{F}_s$ is equivalent to the following \ac{HS}:

\begin{equation}
\mathcal{F}_s \triangleq \{ \boldsymbol{x} \mapsto ( \lambda_1 \left \langle \boldsymbol{w}_1, \phi(\boldsymbol{x}) \right \rangle, \cdots, \lambda_T \left \langle \boldsymbol{w}_T, \phi(\boldsymbol{x}) \right \rangle)' : \left \| \boldsymbol{w}_t \right \|^2 \leq R, \boldsymbol{\lambda} \in \Omega_s(\boldsymbol{\lambda}) \}
\label{eq:HS_equivalent_supp}
\end{equation}

\noindent
According to the same reasoning of Equations (1) and (2) in \cite{Cortes2010}, we know that $\boldsymbol{w}_t = \sum_{i=1}^N \alpha_t^i \phi(\boldsymbol{x}_t^i)$, and the constraint $\left \| \boldsymbol{w}_t\right \|^2 \leq R$ is equivalent to $\boldsymbol{\alpha}_t^{'} \boldsymbol{K}_t \boldsymbol{\alpha}_t \leq R$. Therefore, based on the definition of \ac{ERC} that is given in \eref{eq:rademacher}, we have that

\begin{equation}
\hat{R}(\mathcal{F}_s) = \frac{2}{TN} E_{\sigma} \{\sup_{\boldsymbol{\alpha}_t \in \mathcal{F}_s} \sum_{t=1}^{T} \lambda_t \boldsymbol{\sigma}_t^{'} \boldsymbol{K}_t \boldsymbol{\alpha}_t \}
\label{eq:rademacher_original_supp}
\end{equation}

\noindent 
where $\mathcal{F}_s = \{ \boldsymbol{\alpha}_t \; | \;\boldsymbol{\alpha}_t^{'} \boldsymbol{K}_t \boldsymbol{\alpha}_t \leq \lambda_t^2 R, \forall t;  \boldsymbol{\lambda} \in \Omega_s (\boldsymbol{\lambda}) \}$. To solve the maximization problem with respect to $\boldsymbol{\alpha}_t$, we observe that the $T$ problems are independent and thus can be solved individually. Based on Cauchy-Schwartz inequality, the optimal $\boldsymbol{\alpha}_t$ is achieved when $\boldsymbol{K}_t^{\frac{1}{2}} \boldsymbol{\alpha}_t = c_t \boldsymbol{K}_t^{\frac{1}{2}} \boldsymbol{\sigma}_t$ where $c_t$ is a constant. 

Substituting this result into each of the $T$ maximization problems, we have the following:

%\gcacomment{Do not use ``plug in'', ``plugging it''! It is colloquial English. Use ``substitute in'' or similar. (fixed)}

\begin{equation}
\begin{aligned}
\max_{c_t} & \;\;\;\; c_t \boldsymbol{\sigma}_t^{'} \boldsymbol{K}_t \boldsymbol{\sigma}_t \\
\text{s.t.} & \;\;\;\; c_t^2 \boldsymbol{\sigma}_t^{'} \boldsymbol{K}_t \boldsymbol{\sigma}_t \leq R
\end{aligned}
\label{eq:max_ct_supp}
\end{equation}

\noindent 
Obviously, the optimal $c_t$ is obtained when $c_t = \sqrt{\frac{R}{\boldsymbol{\sigma}_t^{'} \boldsymbol{K}_t \boldsymbol{\sigma}_t}}$. Therefore the \ac{ERC} becomes now

\begin{equation}
\hat{R}(\mathcal{F}_s) = \frac{2}{TN} E_{\sigma} \{\sup_{\boldsymbol{\lambda} \in \Omega_s(\boldsymbol{\lambda})} \sum_{t=1}^{T} \lambda_t \sqrt{\boldsymbol{\sigma}_t^{'} \boldsymbol{K}_t \boldsymbol{\sigma}_t R} \}
\label{eq:rademacher_without_alpha_supp}
\end{equation}

\noindent
Since $s \geq 1$, based on \lemmaref{lemma:simple_max_problem}, it is not difficult to get the solution of the maximization problem with respect to $\boldsymbol{\lambda}$, which gives

\begin{equation}
\begin{aligned}
\hat{R}(\mathcal{F}_s) & = \frac{2\sqrt{R}}{TN} E_{\sigma} \{ [\sum_{t=1}^{T} ( \boldsymbol{\sigma}_t^{'} \boldsymbol{K}_t \boldsymbol{\sigma}_t )^{\frac{s^*}{2}}]^{\frac{1}{s^*}} \} \\
 & = \frac{2\sqrt{R}}{TN} E_{\sigma} \{\left \| \boldsymbol{u} \right \|_{s^*}\}
\end{aligned}
\label{eq:rademacher_after_optimization_supp}
\end{equation} 
\end{proof}

%\gcacomment{It is kind of funny: you elaborate on the easiest optimization problem (involving the Cauchy-Schwartz ineq.) and you just casually hold for granted the solution to the last, more difficult optimization problem. I would at least provide its solution in the general case and then say that I apply it to what I have here.}
%\mycomment{Added one subsection at the beginning of Appendix.}

%%%%%%%%%%%%%%%%%%%%%%%%%%%%%%%%%%%%%%%%%%%%%%
%%%%%%%%%%%%%%%%%%%%%%%%%%%%%%%%%%%%%%%%%%%%%%
\subsection{Proof to \thmref{thm:rademacher_monotonicity_fix_feature}}
\label{sec:Proof_to_rademacher_monotonicity_fix_feature}
\begin{proof}
First note that $\forall s_1 > s_2 \geq 1$, we have that $1 \leq s_1^* < s_2^*$, which means $\left \| \boldsymbol{u} \right \|_{s_1^*} \geq \left \| \boldsymbol{u} \right \|_{s_2^*}$. Based on \eref{eq:rademacher_after_optimization_supp}, we immediately have $\hat{R}(\mathcal{F}_{s_1}) \geq \hat{R}(\mathcal{F}_{s_2})$. This gives the monotonicity of $\hat{R}(\mathcal{F}_s)$ with respect to $s$.
\end{proof}

%%%%%%%%%%%%%%%%%%%%%%%%%%%%%%%%%%%%%%%%%%%%%%
%%%%%%%%%%%%%%%%%%%%%%%%%%%%%%%%%%%%%%%%%%%%%%
\subsection{Proof to \thmref{thm:s_infty_fix_feature}}
\label{sec:Proof_to_s_infty_fix_feature}
\begin{proof}
Similar to the proof to \lemmaref{lemma:rademacher_alternative_fix_feature}, we write the \ac{ERC} of $\tilde{\mathcal{F}}$:

\begin{equation}
\hat{R}(\tilde{\mathcal{F}}) = \frac{2}{TN} E_{\sigma} \{\sup_{\boldsymbol{\alpha}_t \in \tilde{\mathcal{F}}} \sum_{t=1}^{T} \boldsymbol{\sigma}_t^{'} \boldsymbol{K}_t \boldsymbol{\alpha}_t \}
\label{eq:rademacher_f_tilde_supp}
\end{equation}

\noindent
Optimize with respect to $\boldsymbol{\alpha}_t$ gives

\begin{equation}
\hat{R}(\tilde{\mathcal{F}}) = \frac{2\sqrt{R}}{TN} E_{\sigma} \{ \sum_{t=1}^{T} \sqrt{\boldsymbol{\sigma}_t^{'} \boldsymbol{K}_t \boldsymbol{\sigma}_t} \}
\label{eq:rademacher_f_tilde_after_optimization_supp}
\end{equation}

\noindent 
Based on \lemmaref{lemma:rademacher_alternative_fix_feature}, we immediately obtain $\hat{R}(\tilde{\mathcal{F}}) = \hat{R}(\mathcal{F}_{+\infty})$. 
\end{proof}

%%%%%%%%%%%%%%%%%%%%%%%%%%%%%%%%%%%%%%%%%%%%%%
%%%%%%%%%%%%%%%%%%%%%%%%%%%%%%%%%%%%%%%%%%%%%%
\subsection{Proof to \thmref{thm:rademacher_fix_feature}}
\label{sec:Proof_to_rademacher_fix_feature}
\begin{proof}
According to \eref{eq:rademacher_after_optimization_supp} and Jensen's Inequality, we have 

%\gcacomment{I would prefer, if you stuck the eqn number here instead of referring to a proof.}
%\mycomment{Done.}

\begin{equation}
\begin{aligned}
\hat{R}(\mathcal{F}_s) & \leq \frac{2\sqrt{R}}{TN}  ( \sum_{t=1}^{T} E_{\sigma} \{ ( \boldsymbol{\sigma}_t^{'} \boldsymbol{K}_t \boldsymbol{\sigma}_t )^{\frac{s^*}{2}} \} ) ^{\frac{1}{s^*}} \\
& = \frac{2\sqrt{R}}{TN}  ( \sum_{t=1}^{T} E_{\sigma} \{ \| \sum_{i=1}^N \sigma_t^i \phi(\boldsymbol{x}_t^i) \|^{s^*} \} ) ^{\frac{1}{s^*}}
\end{aligned}
\label{eq:rademacher_jensen_supp}
\end{equation}

\noindent
Based on \lemmaref{lemma:k-k}, we have that

%\gcacomment{It would have been nice, if you restated it somewhere. Actually, I have an idea: you can include Kahane's ineq. as well as the statements \& solns to all elementary optimization problems you use in your proofs as a small section at the top of the Appendix and then refer to them from inside the proofs. That would be helpful, because in the way the appendix is right now, while I can follow the proofs, I can't easily verify some steps (it would be time-consuming for me)}
%\mycomment{Please see my comment at the beginning of the Appendix regarding the K.-K. inequality.}

\begin{equation}
\begin{aligned}
\hat{R}(\mathcal{F}_s) & \leq \frac{2\sqrt{R}}{TN}  ( \sum_{t=1}^{T} (s^* \text{tr}(\boldsymbol{K}_t))^{\frac{s^*}{2}} ) ^{\frac{1}{s^*}} \\
& = \frac{2}{TN} \sqrt{Rs^* \| \text{tr}(\boldsymbol{K}_t)_{t=1}^T\|_{\frac{s^*}{2}}}
\end{aligned}
\label{eq:rademacher_khint_kahane_supp}
\end{equation}

%\gcacomment{Explain the meaning of the notation inside the norm for the last equality.}
%\mycomment{Done.}

\noindent
where $\| \text{tr}(\boldsymbol{K}_t)_{t=1}^T\|_{\frac{s^*}{2}}$ denotes the $l_{\frac{s^*}{2}}$-norm of vector $[\text{tr}(\boldsymbol{K}_1), \cdots, \text{tr}(\boldsymbol{K}_T)]'$. Since we assumed that $k(\boldsymbol{x}, \boldsymbol{x}) \leq 1, \forall \boldsymbol{x}$, we have

%\gcacomment{The last assumption is nowhere to be found in the statement of the theorems and, most importantly, in your HS definitions!!!}
%
%\mycomment{I don't understand... This assumption is stated in Theorem 3; could you double check? Also, I don't think we should add it in the HS definition. The HS is supposed to be general, and this assumption is just require the kernel function to be normalized, which is easy to do in practice. In fact, even though it is not normalized, it will only change the bound by a constant factor, which we don't care about too much. Such assumption is for the sake of simpler notation and for the theorem only (if we assume $k(\cdot, \cdot) \leq C$, then we are adding one more letter in the bound with no benefit at all).}
%
%\gcacomment{Reg. assumption: indeed. Also, if you use C, instead of 1, you can see what the effect is of the range of the training data on the bound; but maybe this observation is of low gain. The remaining stuff: if your HS is a RKHS, then I think the constraint that Kij < C or something similar should be included. We can discuss...}

\begin{equation}
\begin{aligned}
\hat{R}(\mathcal{F}_s) & \leq \frac{2}{TN} \sqrt{Rs^* (\sum_{t=1}^T N^{\frac{s^*}{2}})^{\frac{2}{s^*}}} \\
& = \frac{2}{T\sqrt{N}} \sqrt{RT^{\frac{2}{s^*}}s^*}
\end{aligned}
\label{eq:rademacher_kernel_less_than_1_supp}
\end{equation}

%\gcacomment{What is N? the common number of samples for each task?}
%\mycomment{Yes. This is stated at the beginning of Section 2.}

Note that this bound can be further improved for the interval $s \in [1, \rho^*]$. To make this improvement, we first prove that $\hat{R}(\mathcal{F}_s) \leq T^{\frac{1}{s'} - \frac{1}{s}} \hat{R}(\mathcal{F}_{s'})$ for any $s' \geq s \geq 1$. 

\begin{equation}
\begin{aligned}
 & \hat{R}(\mathcal{F}_s)  = \frac{2}{TN} E_{\sigma} \{\sup_{\boldsymbol{\lambda} \succeq \boldsymbol{0}, \| \boldsymbol{\lambda} \|_s \leq 1} \sum_{t=1}^{T} \lambda_t \sqrt{\boldsymbol{\sigma}_t^{'} \boldsymbol{K}_t \boldsymbol{\sigma}_t R} \} \\
 & \leq \frac{2}{TN} E_{\sigma} \{\sup_{\boldsymbol{\lambda} \succeq \boldsymbol{0}, \| \boldsymbol{\lambda} \|_{s'} \leq T^{\frac{1}{s'} - \frac{1}{s}}} \sum_{t=1}^{T} \lambda_t \sqrt{\boldsymbol{\sigma}_t^{'} \boldsymbol{K}_t \boldsymbol{\sigma}_t R} \} \\
 & = \frac{2}{TN} E_{\sigma} \{\sup_{\boldsymbol{\lambda} \succeq \boldsymbol{0}, \| \boldsymbol{\lambda} \|_{s'} \leq 1} \sum_{t=1}^{T} T^{\frac{1}{s'} - \frac{1}{s}} \lambda_t \sqrt{\boldsymbol{\sigma}_t^{'} \boldsymbol{K}_t \boldsymbol{\sigma}_t R} \} \\
 & = T^{\frac{1}{s'} - \frac{1}{s}} \hat{R}(\mathcal{F}_{s'})
\end{aligned}
\label{eq:rademacher_relation_regarding_s_supp}
\end{equation}

\noindent
Based on this conclusion, we have that $\forall s \in [1, \rho^*]$, 

\begin{equation}
\begin{aligned}
 & \hat{R}(\mathcal{F}_s)  \leq T^{\frac{1}{\rho^*} - \frac{1}{s}} \hat{R}(\mathcal{F}_{\rho^*}) \\
 & = T^{\frac{1}{\rho^*} - \frac{1}{s}} \frac{2}{TN} \sqrt{2eRN \log T} \\
 & = T^{\frac{1}{\rho^*} -1+1 - \frac{1}{s}} \frac{2}{TN} \sqrt{2eRN \log T} \\
 & = \frac{T^{\frac{1}{s^*}}}{T^{\frac{1}{\rho}}} \frac{2}{TN} \sqrt{2eRN \log T} \\
 & = \frac{T^{\frac{1}{s^*}}}{\sqrt{e}} \frac{2}{TN} \sqrt{2eRN \log T} \\
 & = \frac{2}{T\sqrt{N}} \sqrt{RT^{\frac{2}{s^*}} \rho}
\end{aligned}
\label{eq:rademacher_improved_supp}
\end{equation}

\noindent
Note that this is always less than $\frac{2}{T\sqrt{N}} \sqrt{RT^{\frac{2}{s^*}}s^*}$ that is given in (\ref{eq:rademacher_kernel_less_than_1_supp}); $\rho^*$ is the global minimizer of the expression in (\ref{eq:rademacher_kernel_less_than_1_supp}) as a function of $s$. In summary, we have $\hat{R}(\mathcal{F}_s) \leq \frac{2}{T\sqrt{N}} \sqrt{RT^{\frac{2}{s^*}} \rho}$ when $s \in [1, \rho^*]$, and $\hat{R}(\mathcal{F}_s) \leq \frac{2}{T\sqrt{N}} \sqrt{RT^{\frac{2}{s^*}}s^*}$ when $s > \rho^*$.
\end{proof}

%%%%%%%%%%%%%%%%%%%%%%%%%%%%%%%%%%%%%%%%%%%%%%
%%%%%%%%%%%%%%%%%%%%%%%%%%%%%%%%%%%%%%%%%%%%%%
\subsection{Proof to \lemmaref{lemma:rademacher_alternative_learn_feature}}
\label{sec:Proof_to_rademacher_alternative_learn_feature}
\begin{proof}
Define $\boldsymbol{K}_t \triangleq \sum_{m=1}^M \theta_m \boldsymbol{K}_t^m, \forall t$, then we can write

\begin{equation}
\hat{R}(\mathcal{F}_{s, r}) = \frac{2}{TN} E_{\sigma} \{\sup_{\boldsymbol{\alpha}_t \in \mathcal{F}_{s, r}} \sum_{t=1}^{T} \lambda_t \boldsymbol{\sigma}_t^{'} \boldsymbol{K}_t \boldsymbol{\alpha}_t \}
\label{eq:rademacher_learn_feature_original_supp}
\end{equation}

\noindent
where $\mathcal{F}_s = \{ \boldsymbol{\alpha}_t \; | \;\boldsymbol{\alpha}_t^{'} \boldsymbol{K}_t \boldsymbol{\alpha}_t \leq \lambda_t^2 R, \forall t;  \boldsymbol{\lambda} \in \Omega_s (\boldsymbol{\lambda}); \boldsymbol{\theta} \in \Omega_r(\boldsymbol{\theta}) \}$. Then using the similar proof of \lemmaref{lemma:rademacher_alternative_fix_feature}, we have that 

\begin{equation}
\begin{aligned}
\hat{R}(\mathcal{F}_{s, r}) & = \frac{2\sqrt{R}}{TN} E_{\sigma} \{ \sup_{\boldsymbol{\theta} \in \Omega_r(\boldsymbol{\theta})}[\sum_{t=1}^{T} ( \boldsymbol{\sigma}_t^{'} \boldsymbol{K}_t \boldsymbol{\sigma}_t )^{\frac{s^*}{2}}]^{\frac{1}{s^*}} \} \\
 & = \frac{2\sqrt{R}}{TN} E_{\sigma} \{\sup_{\boldsymbol{\theta} \in \Omega_r(\boldsymbol{\theta})} \sum_{t=1}^T (\boldsymbol{\theta}' \boldsymbol{u}_t)^{\frac{s^*}{2}}\}^{\frac{1}{s^*}}
\end{aligned}
\label{eq:rademacher_learn_feature after_optimization_supp}
\end{equation} 

\noindent
This gives the first equation in \eref{eq:rademacher_alternative_learn_feature_1}. To prove the second equation, we simply optimize (\ref{eq:rademacher_learn_feature after_optimization_supp}) with respect to $\boldsymbol{\theta}$, which directly gives the result.
\end{proof}
%
%\gcacomment{Make a final sweep to the document to see if the references to equations are all correct. So far, I caught two mistakes of referring to the wrong equation.}
%
%\mycomment{I'll do it in the final round.}

%%%%%%%%%%%%%%%%%%%%%%%%%%%%%%%%%%%%%%%%%%%%%%
%%%%%%%%%%%%%%%%%%%%%%%%%%%%%%%%%%%%%%%%%%%%%%
\subsection{Proof to \thmref{thm:rademacher_monotonicity_learn_feature}}
\label{sec:Proof_to_rademacher_monotonicity_learn_feature}
\begin{proof}
Consider \eref{eq:rademacher_alternative_learn_feature_1} and let $g(\boldsymbol{\lambda}) \triangleq \sup_{\boldsymbol{\alpha} \in \Omega(\boldsymbol{\alpha})} \| \boldsymbol{v} \|_{r^*}$. Then 

\begin{equation}
\hat{R}(\mathcal{F}_{s, r}) = \frac{2}{TN} E_{\sigma} \{ \sup_{\boldsymbol{\lambda} \in \Omega_s(\boldsymbol{\lambda})} g(\boldsymbol{\lambda})\} 
\label{eq:rademacher_learn_feature_optimize_lambda_supp}
\end{equation}

Note that $\forall 1 \leq s_1 < s_2 $, we have the relation $\Omega_{s_1}(\boldsymbol{\lambda}) \subseteq \Omega_{s_2}(\boldsymbol{\lambda})$. Therefore, let $\hat{\boldsymbol{\lambda}}_1$, $\hat{\boldsymbol{\lambda}}_2$ be the solution of problems $\sup_{\boldsymbol{\lambda} \in \Omega_{s_1}(\boldsymbol{\lambda})} g(\boldsymbol{\lambda})$ and $\sup_{\boldsymbol{\lambda} \in \Omega_{s_2}(\boldsymbol{\lambda})} g(\boldsymbol{\lambda})$ correspondingly, we must have $g(\boldsymbol{\hat{\boldsymbol{\lambda}}}_1) \leq g(\boldsymbol{\hat{\boldsymbol{\lambda}}}_2)$. This directly implies $\hat{R}(\mathcal{F}_{s_1, r}) \leq \hat{R}(\mathcal{F}_{s_2, r})$.
\end{proof}

%%%%%%%%%%%%%%%%%%%%%%%%%%%%%%%%%%%%%%%%%%%%%%
%%%%%%%%%%%%%%%%%%%%%%%%%%%%%%%%%%%%%%%%%%%%%%
\subsection{Proof to \thmref{thm:s_infty_learn_feature}}
\label{sec:Proof_to_s_infty_learn_feature}
\begin{proof}
Define $\boldsymbol{K}_t \triangleq \sum_{m=1}^M \theta_m \boldsymbol{K}_t^m, \forall t$, then

\begin{equation}
\hat{R}(\tilde{\mathcal{F}}_r) = \frac{2}{TN} E_{\sigma} \{\sup_{\boldsymbol{\alpha}_t \in \tilde{\mathcal{F}}} \sum_{t=1}^{T} \boldsymbol{\sigma}_t^{'} \boldsymbol{K}_t \boldsymbol{\alpha}_t \}
\label{eq:rademacher_f_tilde_learn_feature_supp}
\end{equation}

\noindent
Fix $\boldsymbol{\theta}$ and optimize with respect to $\boldsymbol{\alpha}_t$ gives

\begin{equation}
\hat{R}(\tilde{\mathcal{F}}_r) = \frac{2}{TN} \sqrt{R} E_{\sigma} \{\sup_{\boldsymbol{\theta} \in \Omega_r(\boldsymbol{\theta})} \sum_{t=1}^T \sqrt{\boldsymbol{\theta}' \boldsymbol{u}_t}\}
\label{eq:rademacher_f_tilde_after_optimization_learn_feature_supp}
\end{equation}

\noindent
Based on \eref{eq:rademacher_alternative_learn_feature_1}, we immediately obtain $\hat{R}(\tilde{\mathcal{F}}_r) = \hat{R}(\mathcal{F}_{+\infty, r})$.
\end{proof}

%%%%%%%%%%%%%%%%%%%%%%%%%%%%%%%%%%%%%%%%%%%%%%
%%%%%%%%%%%%%%%%%%%%%%%%%%%%%%%%%%%%%%%%%%%%%%
\subsection{Proof to \thmref{thm:rademacher_learn_feature}}
\label{sec:Proof_to_rademacher_learn_feature}
\begin{proof}
Based on \eref{eq:rademacher_alternative_learn_feature_1} and H\"{o}lder's Inequality, let $c \triangleq \max\{ 0, \frac{1}{r^*} - \frac{2}{s^*}\}$ we have that

\begin{equation}
\begin{aligned}
\hat{R}(\mathcal{F}_{s, r}) & \leq \frac{2}{TN} \sqrt{R} E_{\sigma} \{\sup_{\boldsymbol{\theta} \in \Omega_r(\boldsymbol{\theta})} \sum_{t=1}^T (\|\boldsymbol{\theta}\|_r \|\boldsymbol{u}_t\|_{r*})^{\frac{s^*}{2}}\}^{\frac{1}{s^*}} \\
& = \frac{2}{TN} \sqrt{R} E_{\sigma} \{ \sum_{t=1}^T \|\boldsymbol{u}_t\|_{r*}^{\frac{s^*}{2}}\}^{\frac{1}{s^*}} \\
& \leq \frac{2}{TN} \sqrt{R M^c} E_{\sigma} \{ \sum_{t=1}^T \|\boldsymbol{u}_t\|_{\frac{s^*}{2}}^{\frac{s^*}{2}}\}^{\frac{1}{s^*}}
\end{aligned}
\label{eq:rademacher_learn_feature_holder_supp}
\end{equation}

\noindent
Applying Jensen's Inequality, we have that

\begin{equation}
\begin{aligned}
\hat{R}(\mathcal{F}_{s, r}) & \leq \frac{2}{TN} \sqrt{R M^c} (\sum_{t,m=1}^{T,M} E_{\sigma}\{(u_t^m)^{\frac{s^*}{2}}\})^{\frac{1}{s^*}} \\
& = \frac{2}{TN} \sqrt{R M^c} (\sum_{t,m=1}^{T,M} E_{\sigma} \| \sum_{i=1}^N  \sigma_t^i \phi_m(\boldsymbol{x}_t^i)\|^{s^*})^{\frac{1}{s^*}}
\end{aligned}
\label{eq:rademacher_learn_feature_jensen_supp}
\end{equation}

\noindent
Using \lemmaref{lemma:k-k}, we have that

\begin{equation}
\hat{R}(\mathcal{F}_{s, r}) \leq \frac{2}{TN} \sqrt{R M^c s^*} (\sum_{t,m=1}^{T,M} (\text{tr}(\boldsymbol{K}_t^m))^{\frac{s^*}{2}} )^{\frac{1}{s^*}}
\label{eq:rademacher_learn_feature_khin_kahane_supp}
\end{equation}

\noindent
Since we assume that $k_m(\boldsymbol{x}, \boldsymbol{x}) \leq 1, \forall m, \boldsymbol{x}$, we have that

\begin{equation}
\hat{R}(\mathcal{F}_{s, r}) \leq \frac{2}{T\sqrt{N}} \sqrt{R s^* T^{\frac{2}{s^*}} M^{\max \{ \frac{1}{r^*}, \frac{2}{s^*} \}}}
\label{eq:rademacher_learn_feature_trace_supp}
\end{equation}
\end{proof}

%%%%%%%%%%%%%%%%%%%%%%%%%%%%%%%%%%%%%%%%%%%%%%
%%%%%%%%%%%%%%%%%%%%%%%%%%%%%%%%%%%%%%%%%%%%%%
\subsection{Proof to \colref{col:rademacher_learn_feature_r}}
\label{sec:Proof_to_rademacher_learn_feature_r}
\begin{proof}
First, by following the same proof of $\hat{R}(\mathcal{F}_s) \leq T^{\frac{1}{s'} - \frac{1}{s}} \hat{R}(\mathcal{F}_{s'})$ for any $s' \geq s \geq 1$, we can directly obtain the conclusion that $\hat{R}(\mathcal{F}_{s, r}) \leq T^{\frac{1}{s'} - \frac{1}{s}} \hat{R}(\mathcal{F}_{s', r})$ for any $s' \geq s \geq 1$.

When $r^* \leq \log T$ and $s \in [1, \rho^*]$, where $\rho = 2 \log T$, we have that

\begin{equation}
\begin{aligned}
\hat{R}(\mathcal{F}_{s, r}) & \leq T^{\frac{1}{\rho^*} - \frac{1}{s}} \hat{R}(\mathcal{F}_{\rho^*, r}) \\
& = \frac{2T^{\frac{1}{s^*}}}{T\sqrt{Ne}} \sqrt{2eR  M^{\frac{1}{r^*}}\log T } \\
& = \frac{2}{T\sqrt{N}} \sqrt{R T^{\frac{2}{s^*}} \rho M^{\frac{1}{r^*}}}
\end{aligned}
\label{eq:rademacher_improved_learn_feature_1_supp}
\end{equation}

\noindent
When $s > \rho^*$, obviously, we have $\hat{R}(\mathcal{F}_{s, r}) \leq \frac{2}{T\sqrt{N}} \sqrt{R T^{\frac{2}{s^*}} s^* M^{\frac{1}{r^*}}}$

Similarly, for $r^* \geq \log MT$ and $s \in [1, \rho^*]$, where $\rho = 2 \log MT$, we have that

\begin{equation}
\begin{aligned}
\hat{R}(\mathcal{F}_{s, r}) & \leq T^{\frac{1}{\rho^*} - \frac{1}{s}} \hat{R}(\mathcal{F}_{\rho^*, r}) \\
& = \frac{T^{\frac{1}{s^*}}}{\sqrt{T^{\frac{1}{\log MT}}}} \frac{2}{T\sqrt{N}} \sqrt{2Re \log MT} \\
& = \frac{2}{T\sqrt{N}} \sqrt{2R T^{\frac{2}{s^*}} M^{\frac{1}{\log MT}} \log MT} \\
& = \frac{2}{T\sqrt{N}} \sqrt{R T^{\frac{2}{s^*}} \rho M^{\frac{2}{\rho}}}
\end{aligned}
\label{eq:rademacher_improved_learn_feature_2_supp}
\end{equation}

\noindent
When $s > \rho^*$, obviously, we have $\hat{R}(\mathcal{F}_{s, r}) \leq \frac{2}{T\sqrt{N}} \sqrt{R T^{\frac{2}{s^*}} s^* M^{\frac{2}{s^*}}}$
\end{proof}

%%%%%%%%%%%%%%%%%%%%%%%%%%%%%%%%%%%%%%%%%%%%%%
%%%%%%%%%%%%%%%%%%%%%%%%%%%%%%%%%%%%%%%%%%%%%%
\subsection{Proof to \thmref{thm:group_lasso}}
\label{sec:Proof_to_group_lasso}
\begin{proof}

We have already show that the \ac{ERC} of $\mathcal{F}_s$ is 

\begin{equation}
\hat{R}(\mathcal{F}_s) = \frac{2}{TN} E_{\sigma} \{ \sup_{\boldsymbol{\alpha}, \boldsymbol{\lambda}} \sum_{t=1}^T \boldsymbol{\sigma}_t' \boldsymbol{K}_t \boldsymbol{\alpha}_t \}
\label{eq:proof_gl_1}
\end{equation}

\noindent
It is not difficult to see that optimizing the following problem 

\begin{equation}
\begin{aligned}
\sup_{\boldsymbol{\alpha}, \boldsymbol{\lambda}} & \sum_{t=1}^T \boldsymbol{\sigma}_t' \boldsymbol{K}_t \boldsymbol{\alpha}_t \\
\textit{s.t.}\; & \boldsymbol{\alpha}_t' \boldsymbol{K}_t \boldsymbol{\alpha}_t \leq \lambda_t^2 R \\
 & \| \boldsymbol{\lambda} \|_s \leq 1
\end{aligned}
\label{eq:proof_gl_2}
\end{equation}

\noindent
with respect to $\boldsymbol{\alpha}_t$ must achieves its optimum at the boundary, \ie, the optimal $\boldsymbol{\alpha}_t$ must satisfy $\boldsymbol{\alpha}_t' \boldsymbol{K} \boldsymbol{\alpha}_t = \lambda_t^2 R$. Therefore, \pref{eq:proof_gl_2} can be re-written as

\begin{equation}
\begin{aligned}
\sup_{\boldsymbol{\alpha}, \boldsymbol{\lambda}} & \sum_{t=1}^T \boldsymbol{\sigma}_t' \boldsymbol{K}_t \boldsymbol{\alpha}_t \\
\textit{s.t.}\; & \boldsymbol{\alpha}_t' \boldsymbol{K}_t \boldsymbol{\alpha}_t = \lambda_t^2 R \\
 & \| \boldsymbol{\lambda} \|_s \leq 1
\end{aligned}
\label{eq:proof_gl_3}
\end{equation}

\noindent
Substituting the first constraint into the second one directly leads to the result. The proof regarding to $\mathcal{F}_{s, r}$ is similar, and therefore we omit it.
\end{proof}

\end{document}